\documentclass[11pt]{article}

\usepackage[preprint]{acl}

\usepackage{times}
\usepackage{latexsym}
\usepackage[T1]{fontenc}
\usepackage[utf8]{inputenc}
\usepackage{microtype}
\usepackage{inconsolata}
\usepackage{graphicx}
\usepackage{duckuments}

\usepackage{url}
\usepackage{booktabs}
\usepackage{amsfonts}
\usepackage{amssymb}
\usepackage{amsmath}
\usepackage{nicefrac}
\usepackage{mathtools,bbm}
\usepackage[dvipsnames]{xcolor}
\usepackage{multirow}
\usepackage{caption}
\usepackage{subcaption}
\usepackage{makecell}
\usepackage{wrapfig}
\usepackage{cleveref}
\usepackage{textcomp}
\usepackage{gensymb}
\usepackage{marginnote}
\usepackage{colortbl}
\usepackage{enumitem}
\usepackage{pifont}
\usepackage{lipsum}
\usepackage{pifont}
\usepackage{fontawesome}
\usepackage{fvextra}

\usepackage[colorinlistoftodos, textsize=small]{todonotes}

\usepackage{tikz}
\usetikzlibrary{arrows.meta, positioning, shapes.geometric, fit, backgrounds, calc}
\usepackage{algorithm}
\usepackage{algpseudocode}

\crefname{equation}{}{}

\definecolor{BrickRed}{rgb}{0.6,0,0}
\definecolor{RoyalBlue}{rgb}{0,0,0.8}
\definecolor{Tdgreen}{rgb}{0,0.4,0.7}
\definecolor{darkblue}{rgb}{0,0.08,0.45}
\hypersetup{citecolor=darkblue, linkcolor=darkblue}

\title{LLMs Get Lost in Evolving User Intent}

\author{
Jihoon Tack, Philippe Laban, Jennifer Neville\\
\IfFileExists{logos/microsoft.png}{\raisebox{-0.25ex}{\includegraphics[height=2ex]{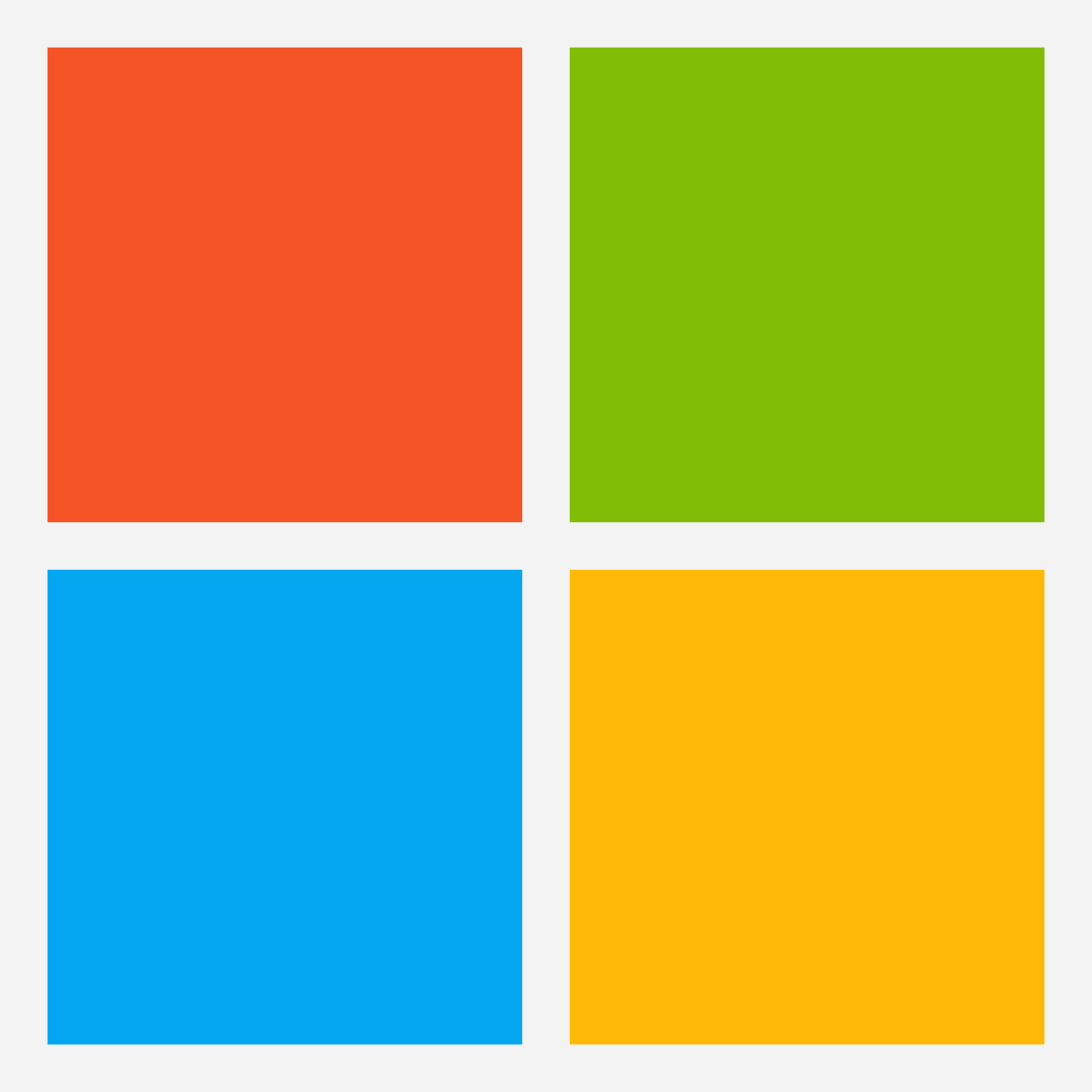}}\hspace{0.25pt}}{} Microsoft Research\\
\texttt{\{jihoontack,plaban,jenneville\}@microsoft.com}\\
\href{https://github.com/microsoft/evolving-intent/}{\color{black}\faGithub~\texttt{microsoft/evolving-intent/}}
}

\begin{document}
\maketitle

\begin{abstract}
As LLMs become more capable, they are increasingly deployed as collaborative agents, taking on user-delegated tasks through iterative interaction. Yet genuine interaction is inherently dynamic: users rarely specify their intent upfront, instead disclosing, revising, and reshaping it as the conversation unfolds. Despite this, LLMs are still predominantly evaluated or trained in single-turn, fully-specified settings, leaving open a fundamental question: how well do LLMs track and act on user intent as it evolves over the course of a conversation? To study this, we introduce a framework that transforms static, single-turn tasks into dynamic multi-turn conversations in which the user's intent evolves across turns—incrementally revealed, revised, and at times redirected mid-conversation—while preserving each task's original evaluation protocol, enabling existing benchmarks to be reused as controlled testbeds without new annotation. Across multiple tasks, we surface a consistent phenomenon: strong static-setting performance does not transfer to the evolving-intent setting, with substantial drops across model families. Our findings point to a fundamental gap: today's LLMs do not yet faithfully track and act on the user's evolving intent, a capability invisible to static evaluation yet critical for future collaborative agents.
\end{abstract}

\section{Introduction}
\label{sec:introduction}

Recent advances in LLMs have driven a shift from passive question-answering chatbots toward LLM-based agents that collaborate with users on long-horizon, real-world tasks such as vibe coding \citep{yang2024swe}, deep research \citep{wei2025browsecomp}, and iterative document editing \citep{laban2026llmsb}. Such collaboration places a new demand on agent capabilities: it must track and act on user intent as it is disclosed, revised, and reshaped throughout the conversation \citep{belkin1982ask, laban2026llmsa}, rather than fully specified upfront. Yet most evaluations of LLMs are still conducted in a single-turn setting \citep{patwardhan2025gdpval, ullrich2026openapps}, where the task is fully specified in a single user turn, leaving open whether current agents can support how users actually interact.

\begin{figure}[t]
    \centering
    \includegraphics[width=\linewidth]{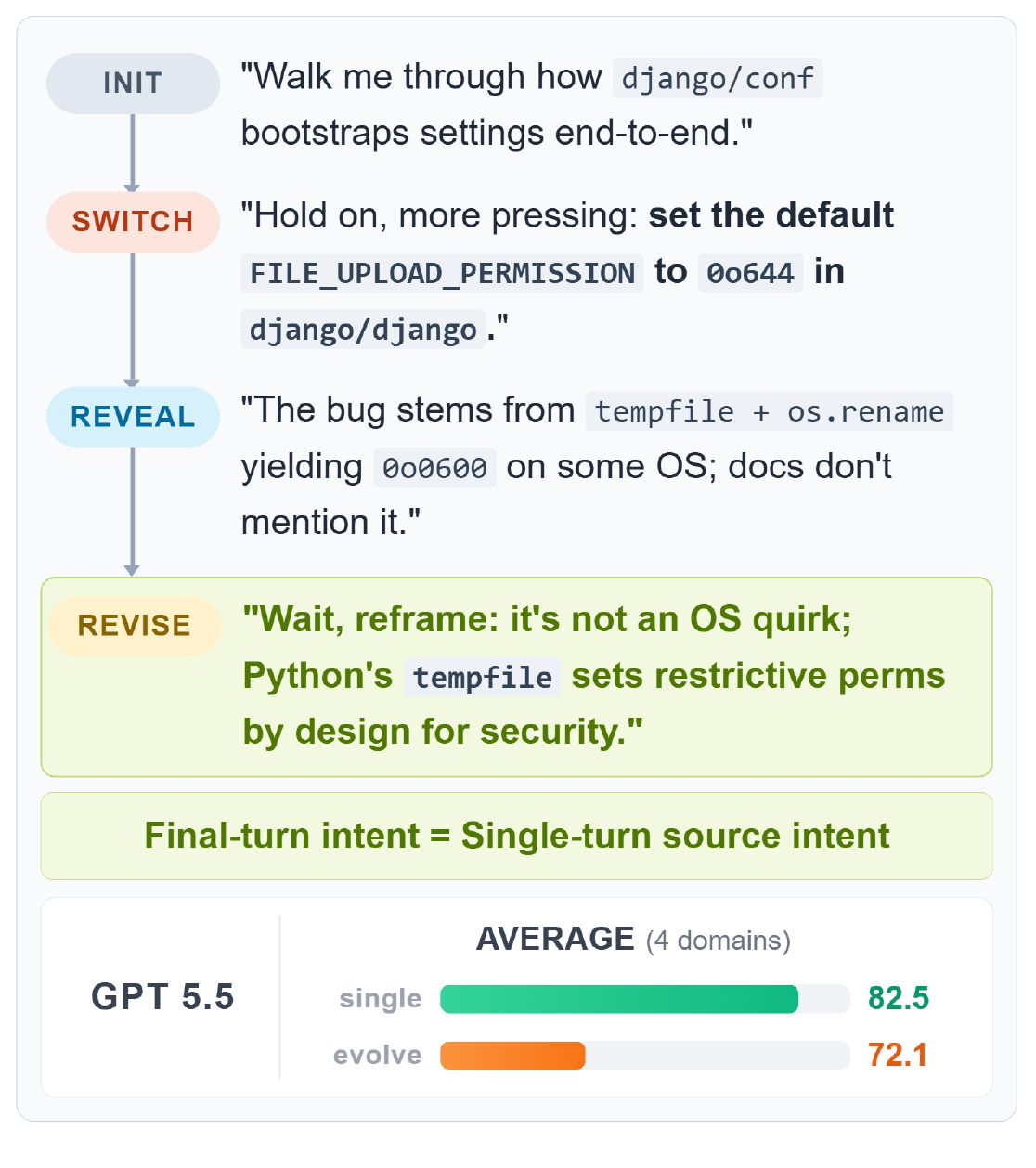}
    \vspace{-0.4in}
    \caption{\textbf{LLMs get lost in evolving user intent.} Utilizing a single-turn instance (e.g., SWE-Bench; \citealp{jimenez2024swe}), we simulate a multi-turn conversation with evolving user intents while preserving the original ground-truth evaluation. Even frontier models degrade substantially after only 6 intent transitions.}
    \label{fig:teaser}
    \vspace{-0.12in}
\end{figure}

To address this gap, a growing body of multi-turn benchmarks has moved beyond static, single-turn evaluation \citep{zheng2023judging,kwan2024mt,wang2024mint,zhou2025sweet}. However, several limitations still prevent them from serving as a scalable evaluation that captures human behavior:(i) rewards are often based on LLM-as-judge rather than verifiable answers, (ii) user turns tend to be short-horizon, and (iii) the user side offers limited controllability over how responses unfold across turns, largely confining user behavior to incremental information disclosure, thereby overlooking the broader dynamics of real users. As a result, they still fall short of providing a scalable and fully verifiable setting for studying how agents track evolving user intent over long interactions.

To tackle this, rather than constructing multi-turn environments from scratch, we revisit static, single-turn benchmarks as a starting point. Despite their stateless nature, they already provide what is costly to obtain in multi-turn environments, i.e., scalable and verifiable supervision, at a scale that hand-authored multi-turn benchmarks cannot match. This raises a natural question: can such static, single-turn tasks be lifted into dynamic multi-turn interactions that capture evolving user intent, while preserving their verifiability?

We propose a framework that converts any verifiable single-turn dataset into a multi-turn environment with evolving user intent, while preserving automatic verifiability against the original answer. We model this evolution through three forms of conversational dynamics: the user may disclose task information incrementally over turns (under-specification), revise previously stated details (revision), or pivot to a related task after completing the current one, where the new task retains part of the previous task context (task switching).

Our key idea is to anchor the user’s intent extracted from source single-turn data at the final turn, and retrospectively construct plausible preceding situations that lead to this anchor. By preserving the source intent at the last turn, the agent’s final action remains directly verifiable using the source dataset’s verifier, without additional annotation or LLM-based judgment. Meanwhile, the earlier turns can be synthesized under controlled intent dynamics, such as progressive revelation, revision, and task switching, enabling scalable long-horizon multi-turn interactions with varying complexity.

We demonstrate the versatility of our framework by transforming benchmarks across multiple domains, including math, text-to-SQL, search, and coding, into dynamic multi-turn environments, and evaluating both frontier and open-source LLMs. Models that excel in the original single-turn setting degrade substantially once the user's intent becomes dynamic: GPT 5.5, for instance, drops from 99.0\% to 80.5\% on math domain after just 6 intent transitions. A closer analysis points to a key factor: current LLMs struggle to track how the user's intent shifts across turns, a capability that static, single-turn evaluation cannot expose but that is essential for real-world multi-turn use.

\section{Related Works}
\label{sec:related}

\paragraph{Multi-turn evaluation.}
With the rise of agentic LLMs, users increasingly accomplish tasks through multi-turn interactions rather than a single fully-specified query \citep{herlihy2024overcoming,baumann2026swe}. Unlike the single-turn, multi-turn interaction induces a combinatorially large space of trajectories that depends on user behavior, significantly broadening the evaluation surface \citep{dou2025simulatorarena}. To reflect this, a growing number of benchmarks have been proposed to measure various multi-turn abilities of LLM agents \citep{zheng2023judging,kwan2024mt,wang2024mint}. However, building such benchmarks at scale incurs substantial annotation cost, since every new task distribution requires fresh conversations and labels, and static conversations offer little control over user behavior, making it hard to systematically probe specific interaction patterns. Moreover, the divergent nature of multi-turn trajectories makes ground truth hard to define, forcing existing benchmarks to use LLM-as-judge protocols whose reliability remains an open question \citep{tang2025learning}.

\begin{figure*}[t]
    \centering\includegraphics[width=\linewidth]{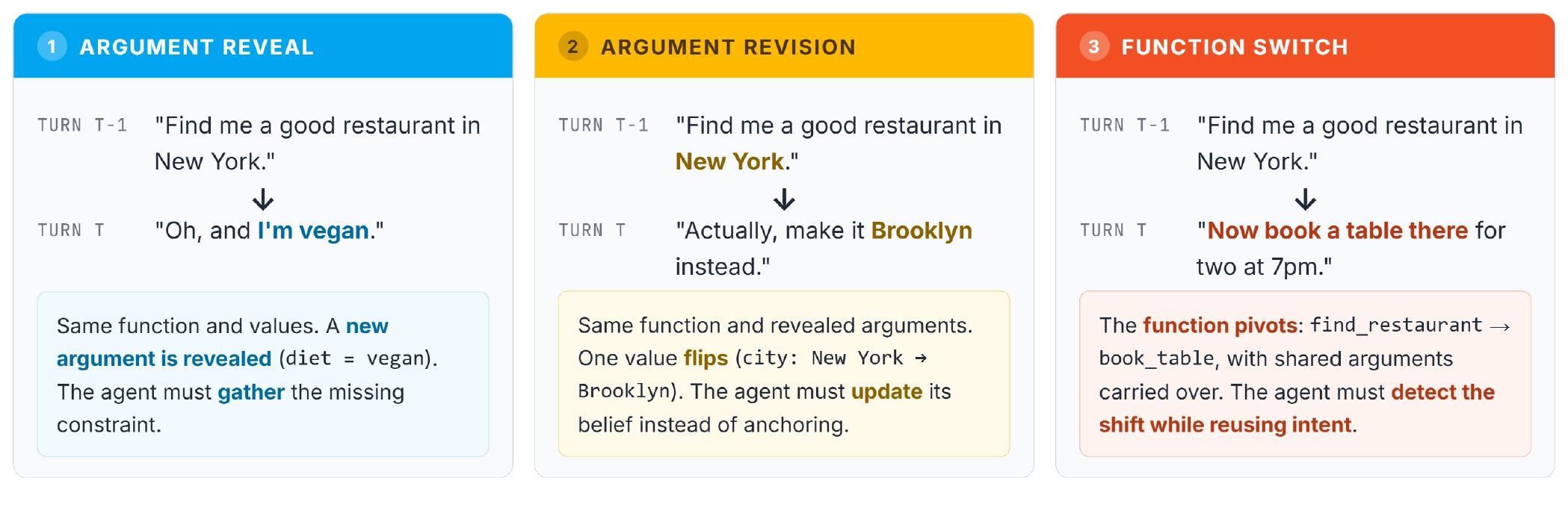}
    \vspace{-0.25in}
    \caption{\textbf{Illustration of three types of intent state transitions.}: \emph{argument reveal}, \emph{argument revision}, and \emph{function switch}. All three start from $\mathcal{I}_{t-1}\!:\,f_{t-1}=\texttt{search\_restaurant},\,c_1=\texttt{New York}$, and at turn $t$ either disclose $c_2=\texttt{vegan}$, update $c_1\!\to\!\texttt{Brooklyn}$, or switch $f_t$ to $\texttt{book\_table}$ with shared arguments carried over.}
    \label{fig:transition-example}
    \vspace{-0.05in}
\end{figure*}

\paragraph{User simulation environment.} 
To this end, recent multi-turn benchmarks increasingly embed simulated user to study how agents handle more realistic interaction \citep{qian2025userrl,zhou2026tom}. One line targets under-specification, approximating it by transforming fully-specified single-turn problems into vague prompts \citep{sun2025training} or by sharding information across turns \citep{laban2026llmsa}. Such constructions cover only a single axis of intent dynamics and are bounded by the source problem's information content, limiting scalability to long-horizon interaction. A separate line instead designs simulated environments with LLM-prompted users \citep{yao2025tau,qian2025userbench,barres2026tau}, which requires per-domain manual curation (e.g., schemas) that may not transfer across task distributions. To address these limitations, we propose a scalable procedure that turns single-turn, fully-specified tasks into long-horizon, complex interactions while preserving a verifiable evaluation signal.

\section{Formulating Evolving User Intent}
\label{sec:method}

We formalize user intent as a controllable, structured state with transition dynamics governing its evolution across turns. Unlike prompting-based LLM simulators \citep{naous2026flipping}, our structured formulation enables precise control over what the user reveals. Moreover, our dynamic transitions go beyond existing simulators that keep users in a fixed under-specified state \citep{sun2025training}, capturing key dynamics such as information revision and task pivoting \citep{baumann2026swe}. This requires the agent to track and revise its understanding of the user's intent, rather than merely accumulating information across turns.

\paragraph{Controllable user intent.}
Concretely, we define a user's intent at turn $t$ as follows:
\begin{align}
  \mathcal{I}_t = (f_t,\; \mathcal{C}_t,\; \mathcal{C}^{\mathtt{rev}}_t, y_t),
\end{align}
where $f_t$ is the function the user wants to accomplish (e.g., $\mathtt{find\_restaurant}$) and $\mathcal{C}_t = \{c_{1,t}, \ldots, c_{N,t}\}$ collects its arguments (e.g., $\{\mathtt{city}, \mathtt{cuisine}\}$), with $c_{i,t}$ the value at the $i$-th slot of $f_t$. The subset $\mathcal{C}^{\mathtt{rev}}_t \subseteq \mathcal{C}_t$ tracks the arguments already revealed to the LLM agent. The answer $y_t$ depends on the task, e.g., a target answer for math and a unit-test suite for coding.

\paragraph{Intent state transitions.}
We propose three types of transitions (see \Cref{fig:transition-example} for illustrative examples). Note that these types are not mutually exclusive and may co-occur within the same turn.

\begin{enumerate}[itemsep=4pt, parsep=0pt, leftmargin=*]
    
    \item \textit{Argument reveal.} The user discloses a previously unrevealed argument:
    \begin{align}
        f_t = f_{t-1},\;\; \mathcal{C}_t = \mathcal{C}_{t-1},\;\; \mathcal{C}^{\mathtt{rev}}_{t-1} \subsetneq \mathcal{C}^{\mathtt{rev}}_t.
    \end{align}
    The underlying intent is fixed; only the agent's observation grows, and the agent must gather the missing information (e.g., \emph{``find me a good restaurant in New York''} $\to$ \emph{``I am vegan''}).

    \item \textit{Argument revision.} The user keeps the task and the same set of revealed slots, but changes the value of at least one revealed argument:
    \begin{align}
    \begin{aligned}
        &f_t = f_{t-1},\;\; |\mathcal{C}^{\mathtt{rev}}_t| = |\mathcal{C}^{\mathtt{rev}}_{t-1}|,\\
    &\exists\, i:\; c_{i,t} \neq c_{i,t-1}.
    \end{aligned}
    \end{align}
    The underlying intent changes ($\mathcal{C}_t \neq \mathcal{C}_{t-1}$), e.g., \emph{New York} $\to$ \emph{Brooklyn}; the agent must update its belief rather than anchor to the previous value.

    \item \textit{Function switch.} The user pivots the task:
    \begin{align}
        f_t \neq f_{t-1}
    \end{align}
    The underlying task itself changes, while shared arguments carry their values across tasks (e.g., \emph{find a restaurant} $\to$ \emph{book it}; $\mathcal{C}_{t-1} \cap \mathcal{C}_{t} \neq \emptyset$); the agent must detect the shift while reusing the relevant intent. 
  
\end{enumerate}

\begin{figure*}[t]
    \centering
    \includegraphics[width=\linewidth]{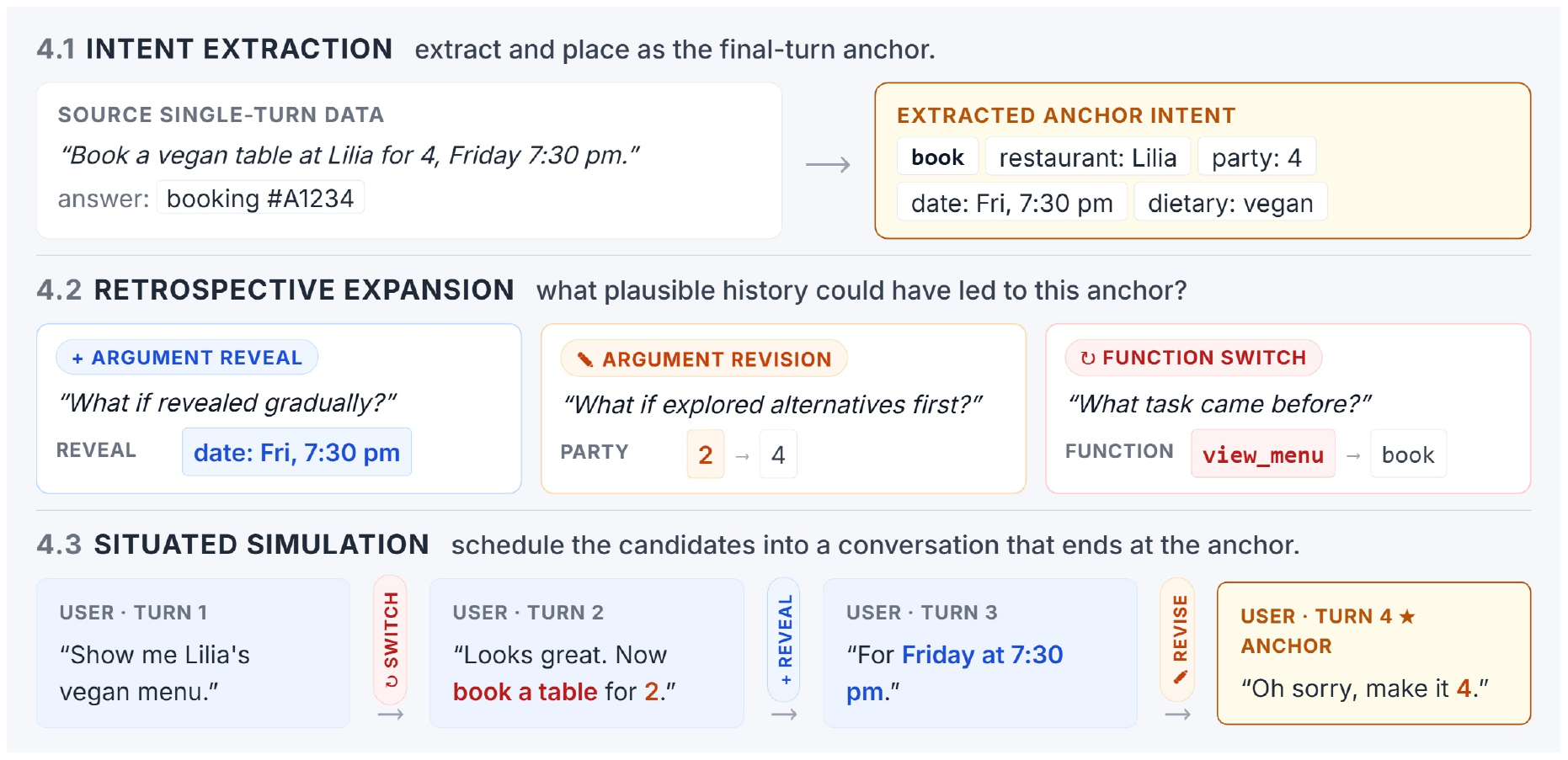}
    \vspace{-0.2in}
    \caption{\textbf{Evolving-intent conversations from single-turn data.} Given a verifiable single turn data, we extract its intent and place it as the final turn anchor. 
    We then synthesize a plausible history by generating counterfactual arguments and expand functions that can would led to this anchor. Finally, we perform a situated simulation so that the conversation ends with the anchor turn. Since the final turn coincides with the source intent by construction, the dataset's original verifier scores the agent's last action without any new annotation.}
    \label{fig:concept}
    \vspace{-0.05in}
\end{figure*}

\section{Simulating Evolving User Intent}

We construct a scalable user simulation framework that supports evolving intent over long-horizon interactions, while remaining automatically verifiable. We first describe how we extract anchor intent from a single-turn verifiable dataset (\Cref{sec:method:anchor}). Then, we retrospectively construct the preceding intents leading to this anchor (\Cref{sec:method:synthesis}), simulating long-horizon multi-turn trajectories that inherit verifiability from the source dataset (\Cref{sec:method:simulation}).

\subsection{Intent Extraction}
\label{sec:method:anchor}

While we aim to simulate users with the proposed intent, the central challenge is \emph{the data}, as per-turn intent and their corresponding labels are prohibitively expensive to collect and annotate at scale. Instead, our idea is to leverage existing single-turn verifiable datasets: from each example, we extract the function, its arguments, and the ground-truth answer, treating this triple as the \emph{anchor intent} for the final conversation turn.

\paragraph{Extracting intent from single-turn data.}
Given a problem-answer pair $(q, y^*)$ from a single-turn dataset, we extract a function and its arguments by prompting an LLM (examples in Appendix~\ref{app:prompts}),
\begin{align}
    (f^*, \mathcal{C}^*) \sim \mathtt{Extract}(\cdot \mid q),
\end{align}
where $f^*$ and $\mathcal{C}^*$ are the extracted source function and argument set, respectively. Our goal is to simulate a realistic conversation $\mathcal{I}_1, \ldots, \mathcal{I}_T$ of turn $T$ that terminates at the source intent:
\begin{align}
  \mathcal{I}_T = (f^*,\; \mathcal{C}^*,\; \mathcal{C}^{\mathtt{rev}}_T = \mathcal{C}^*, y^*),
  \label{eq:anchor}
\end{align}
so that the agent's final action $a_T$ can be scored against $y^*$ by the dataset's verifier.

\subsection{Retrospective Expansion}
\label{sec:method:synthesis}

To expand the conversation that terminates at the anchor intent, we retrospectively synthesize the preceding functions and counterfactual arguments needed for switch and revision. Argument reveal, in contrast, requires no additional synthesis, as the source arguments can simply be distributed across multiple turns.

\paragraph{Counterfactual argument.}
To realize argument revision, we synthesize counterfactual values for source arguments that the user can later revise back to the source. For instance, if the source argument is Brooklyn, the user might first introduce a counterfactual (``Find me a restaurant in New York'') and later revise it to the source (``How about Brooklyn?''). Formally, for each $c_i^* \in \mathcal{C}^*$, we prompt an LLM to generate a counterfactual value,
\begin{align}
    c_i^{\mathtt{cf}} \sim \mathtt{Counterfact}(\cdot \mid c_i^*; \, f^*),
\end{align}
conditioned on the target argument and function.

\paragraph{Predecessor function.}
To realize function switch, we synthesize a predecessor function $f^{\mathtt{pre}}$ such that completing $f^{\mathtt{pre}}$ leads naturally into the source $f^*$. This requires $f^{\mathtt{pre}}$ to share overlapping context with $f^*$, which we enforce by requiring their argument sets to overlap, i.e., $\mathcal{C}^{\mathtt{pre}} \cap \mathcal{C}^* \neq \emptyset$. Formally, the predecessor function (and it's argument set) is as follows,
\begin{align}
    (f^{\mathtt{pre}}, \mathcal{C}^{\mathtt{pre}}) \sim \mathtt{Predecessor}(\cdot \mid f^*, \mathcal{C}^*).
\end{align}
Furthermore, longer predecessor chains are constructed by recursively applying this procedure: each predecessor becomes the new source, yielding a backward chain of the desired depth (e.g., $\mathtt{find\_appointment\_location} \to \mathtt{find\_restaurant} \to \mathtt{book\_restaurant}$). 

\subsection{Situated Simulation}
\label{sec:method:simulation}

We now perform a situated simulation of $T$-turn conversation by (i) \emph{scheduling} which intent transitions occur at each turn and (ii) \emph{rendering} each user turn from the intent update. 

\paragraph{Scheduling intent transitions.}
Given a target turn count $T$ and the budgeted number of each transition type (i.e., function switches, argument reveals, and argument revisions), the scheduler distributes these events across the $T$ turns under a set of consistency rules. Different budget combinations realize different scenarios, ranging from a single function with only progressive reveals to fully combined trajectories with switches and revisions. We highlight the some central rules below; 
\begin{enumerate}[itemsep=3pt, parsep=0pt, leftmargin=*]

    \item \textit{Last turn anchor.} 
    Most critical rule is to end the simulation with the source intent $(f^*, \mathcal{C}^*)$, where all conditions are revealed $\mathcal{C}_T^\mathtt{rev}=\mathcal{C}^*$.

    \item \textit{First turn initialization.}
    The initial turn consists of a function and at least one revealed condition, i.e., $f_0 \neq \texttt{null}$ and $\mathcal{C}_0^{\mathtt{rev}} \neq \emptyset$.

    \item \textit{Within-turn uniqueness.} Each turn carries a single function $f_t$, and the same argument never appears as both $c_i^*$ and $c_i^{\mathtt{cf}}$ within a single $\mathcal{I}_t$ (e.g., a reservation time cannot be both 7:00PM and 1:00PM in the same turn).

    \item \textit{Reveal-before-switch.} A function switch $f_t \to f_{t+1}$ is permitted only when the current task is fully specified, i.e., $\mathcal{C}^{\mathtt{rev}}_t = \mathcal{C}_t$. This ensures the user does not pivot to a new task before the current one is coherent.

    \item \textit{Retrospective rule.} Each predecessor function $f^{\mathtt{pre}}$ must appear before the switch to source function $f^*$, and each counterfactual argument $c_i^{\mathtt{cf}}$ must appear before the revision to its corresponding source argument $c_i^*$.
\end{enumerate}

\paragraph{Rendering user turns.}
For each turn we generate the user response from the intent update $\Delta\mathcal{I}_t := \mathcal{I}_t \setminus \mathcal{I}_{t-1}$ rather than the full intent $\mathcal{I}_t$, mirroring how users typically communicate only what is new. Specifically, a renderer concatenates the function $f_t$ (if updated) with each revealed or revised argument in $\Delta\mathcal{I}_t$ into a natural language. To make the resulting dialogue read naturally, each transition type (function switch, argument reveal, argument revision) is prefixed with a domain-appropriate discourse prefix, e.g., \textit{``Now let's think about this,''} for a function switch and \textit{``Wait, I forgot to mention,''} for a reveal. For settings that require more reactive users, we additionally support an \emph{LLM-based} renderer that rephrases the same content conditioned on the prior agent turn. The full prefix inventory and renderer prompts are presented in \Cref{app:sim_details}.

\begin{table*}[t]
  \centering
  \small
  \caption{\textbf{LLMs get lost in evolving user intent.} Accuracy (\%) across four datasets, comparing fully specified single-turn interactions (\emph{Single}) with evolving-intent interactions (\emph{Evolve}) that compose argument reveal, argument revise, and function switch dynamics. \emph{Evolve} applies each transition type twice, yielding six intent transitions per dialogue (seven turns including the initial turn). Parentheses report the relative change from \emph{Single}, and background color indicates the degradation from each model's \emph{Single} accuracy (saturating at a 50\% relative drop).}
  \label{tab:main}
  \vspace{-0.1in}
  \setlength{\tabcolsep}{5pt}
  \begin{tabular}{l cc cc cc cc}
    \toprule
     & \multicolumn{2}{c}{\textbf{GSM8K}} & \multicolumn{2}{c}{\textbf{BIRD-SQL}} & \multicolumn{2}{c}{\textbf{BrowseComp+}} & \multicolumn{2}{c}{\textbf{SWE-Bench Verif.}} \\
    \cmidrule(lr){2-3} \cmidrule(lr){4-5} \cmidrule(lr){6-7} \cmidrule(lr){8-9}
    Model & Single & Evolve & Single & Evolve & Single & Evolve & Single & Evolve \\
    \midrule
    \IfFileExists{logos/openai.png}{\raisebox{-0.2ex}{\includegraphics[height=2ex]{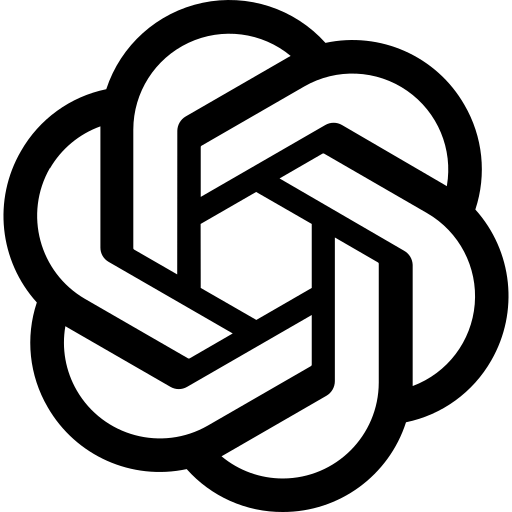}}\hspace{3pt}}{}GPT 5.1 & \cellcolor[HTML]{FFFFFF}\makebox[\widthof{99.0\,{\scriptsize(-99.9)}}][c]{98.0} & \cellcolor[HTML]{FDDFCF}82.0\,{\scriptsize(-16.3)} & \cellcolor[HTML]{FFFFFF}\makebox[\widthof{99.0\,{\scriptsize(-99.9)}}][c]{72.0} & \cellcolor[HTML]{FEF4D8}66.0\,{\scriptsize(-\phantom{0}8.3)} & \cellcolor[HTML]{FFFFFF}\makebox[\widthof{99.0\,{\scriptsize(-99.9)}}][c]{49.0} & \cellcolor[HTML]{E7A0A0}34.0\,{\scriptsize(-30.6)} & \cellcolor[HTML]{FFFFFF}\makebox[\widthof{99.0\,{\scriptsize(-99.9)}}][c]{72.0} & \cellcolor[HTML]{CC8080}\phantom{0}0.0\,{\scriptsize(-100.)} \\
    \IfFileExists{logos/openai.png}{\raisebox{-0.2ex}{\includegraphics[height=2ex]{logos/openai}}\hspace{3pt}}{}GPT 5.2 & \cellcolor[HTML]{FFFFFF}\makebox[\widthof{99.0\,{\scriptsize(-99.9)}}][c]{99.0} & \cellcolor[HTML]{FDDFD0}83.0\,{\scriptsize(-16.2)} & \cellcolor[HTML]{FFFFFF}\makebox[\widthof{99.0\,{\scriptsize(-99.9)}}][c]{77.0} & \cellcolor[HTML]{FACDC9}61.0\,{\scriptsize(-20.8)} & \cellcolor[HTML]{FFFFFF}\makebox[\widthof{99.0\,{\scriptsize(-99.9)}}][c]{53.0} & \cellcolor[HTML]{FEF8EA}50.0\,{\scriptsize(-\phantom{0}5.7)} & \cellcolor[HTML]{FFFFFF}\makebox[\widthof{99.0\,{\scriptsize(-99.9)}}][c]{80.0} & \cellcolor[HTML]{F9C6C6}62.0\,{\scriptsize(-22.5)} \\
    \IfFileExists{logos/openai.png}{\raisebox{-0.2ex}{\includegraphics[height=2ex]{logos/openai}}\hspace{3pt}}{}GPT 5.5 & \cellcolor[HTML]{FFFFFF}\makebox[\widthof{99.0\,{\scriptsize(-99.9)}}][c]{99.0} & \cellcolor[HTML]{FCD5CC}80.5\,{\scriptsize(-18.7)} & \cellcolor[HTML]{FFFFFF}\makebox[\widthof{99.0\,{\scriptsize(-99.9)}}][c]{80.0} & \cellcolor[HTML]{FEEECF}71.0\,{\scriptsize(-11.3)} & \cellcolor[HTML]{FFFFFF}\makebox[\widthof{99.0\,{\scriptsize(-99.9)}}][c]{65.0} & \cellcolor[HTML]{FEEBCF}57.0\,{\scriptsize(-12.3)} & \cellcolor[HTML]{FFFFFF}\makebox[\widthof{99.0\,{\scriptsize(-99.9)}}][c]{86.0} & \cellcolor[HTML]{FEF6E1}80.0\,{\scriptsize(-\phantom{0}7.0)} \\
    \IfFileExists{logos/google.png}{\raisebox{-0.2ex}{\includegraphics[height=2ex]{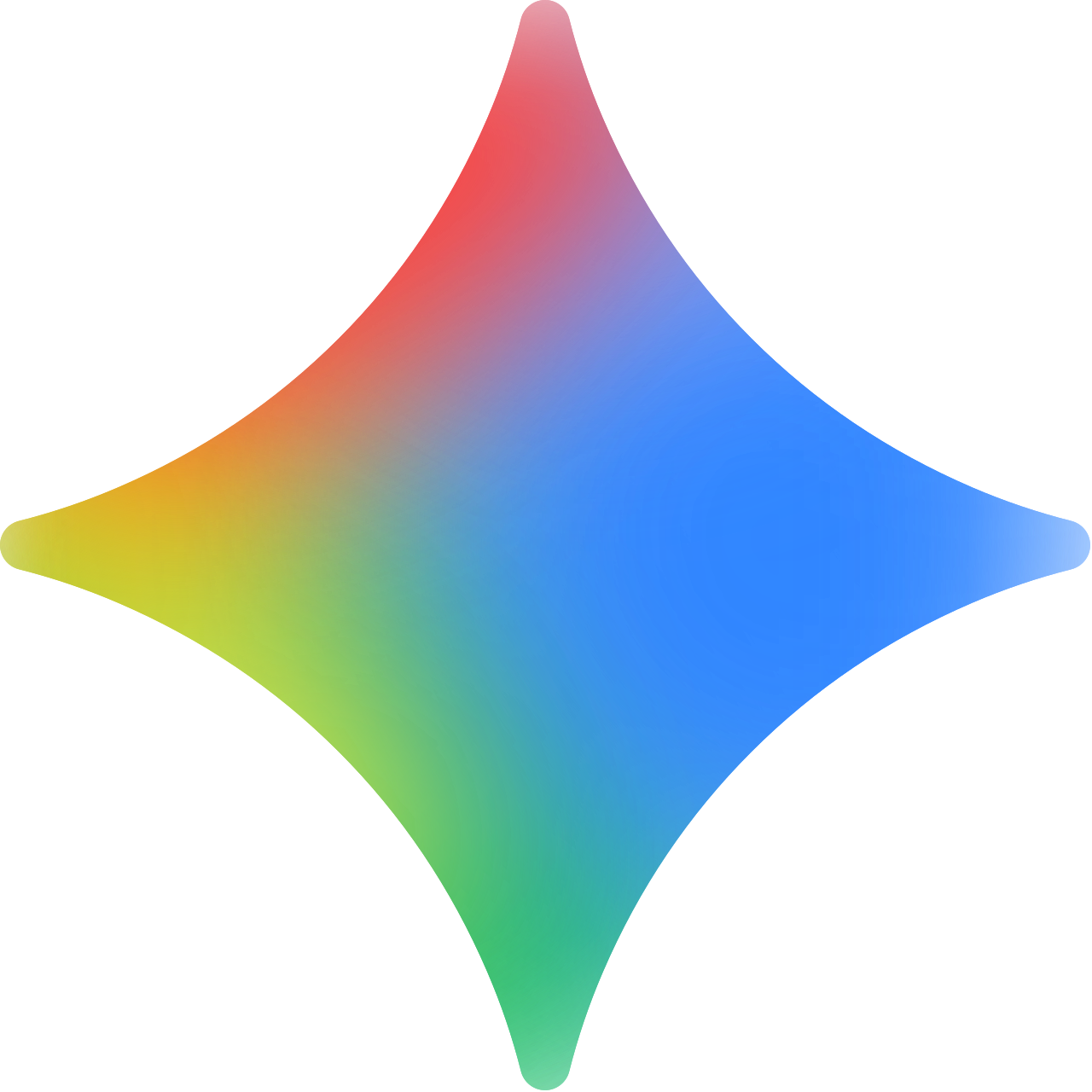}}\hspace{3pt}}{}Gemini 3.1 Pro & \cellcolor[HTML]{FFFFFF}\makebox[\widthof{99.0\,{\scriptsize(-99.9)}}][c]{98.0} & \cellcolor[HTML]{FDDFCF}82.0\,{\scriptsize(-16.3)} & \cellcolor[HTML]{FFFFFF}\makebox[\widthof{99.0\,{\scriptsize(-99.9)}}][c]{75.0} & \cellcolor[HTML]{FEFAF1}72.0\,{\scriptsize(-\phantom{0}4.0)} & \cellcolor[HTML]{FFFFFF}\makebox[\widthof{99.0\,{\scriptsize(-99.9)}}][c]{51.0} & \cellcolor[HTML]{FACAC7}40.0\,{\scriptsize(-21.6)} & \cellcolor[HTML]{FFFFFF}\makebox[\widthof{99.0\,{\scriptsize(-99.9)}}][c]{86.0} & \cellcolor[HTML]{FFFCF7}84.0\,{\scriptsize(-\phantom{0}2.3)} \\
    \IfFileExists{logos/deepseek.png}{\raisebox{-0.2ex}{\includegraphics[height=1.9ex]{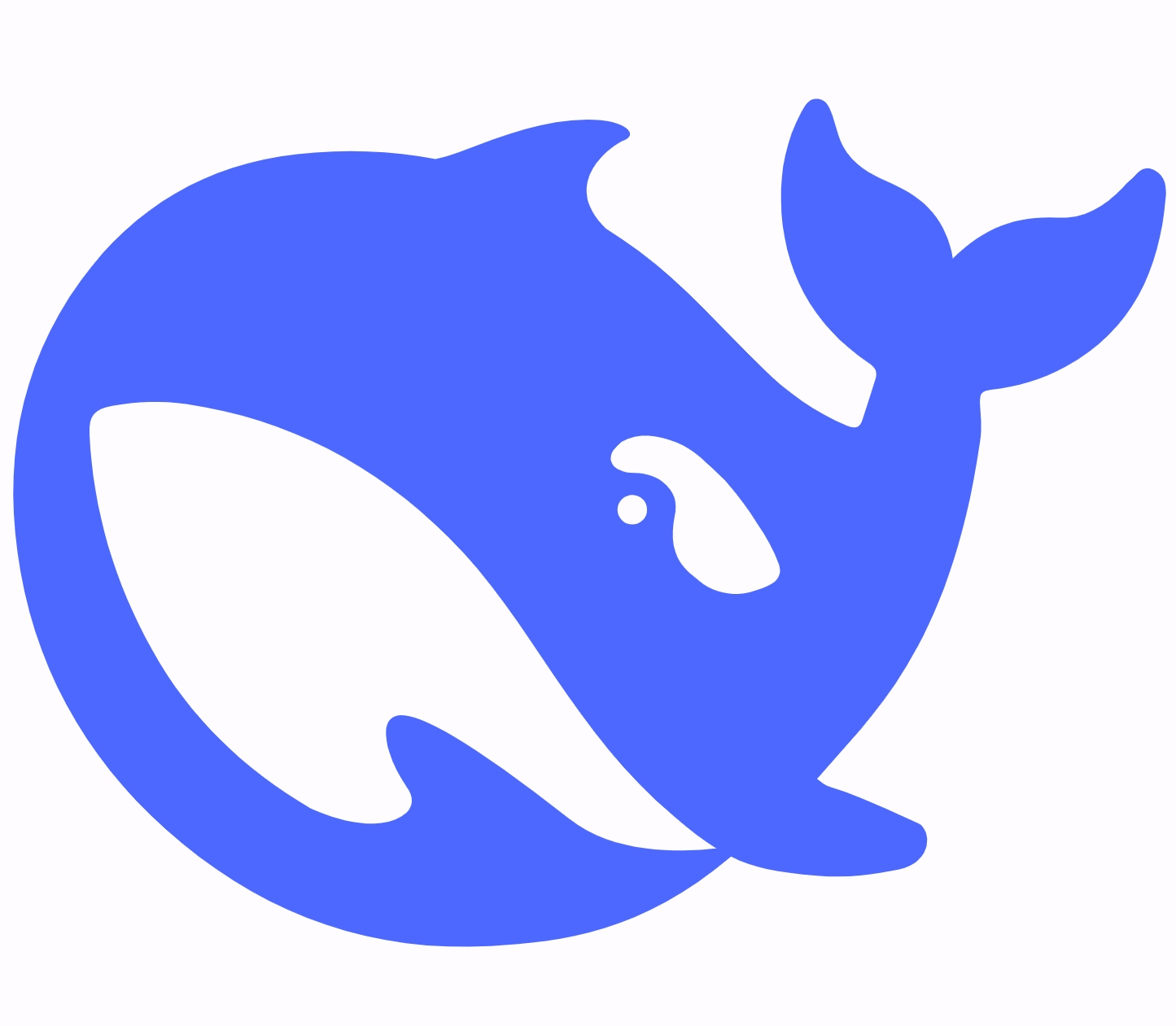}}\hspace{3pt}}{}DeepSeek V3.2 & \cellcolor[HTML]{FFFFFF}\makebox[\widthof{99.0\,{\scriptsize(-99.9)}}][c]{96.5} & \cellcolor[HTML]{FCD5CC}78.5\,{\scriptsize(-18.7)} & \cellcolor[HTML]{FFFFFF}\makebox[\widthof{99.0\,{\scriptsize(-99.9)}}][c]{76.0} & \cellcolor[HTML]{E8A0A0}53.0\,{\scriptsize(-30.3)} & \cellcolor[HTML]{FFFFFF}\makebox[\widthof{99.0\,{\scriptsize(-99.9)}}][c]{36.0} & \cellcolor[HTML]{CC8080}15.0\,{\scriptsize(-58.3)} & \cellcolor[HTML]{FFFFFF}\makebox[\widthof{99.0\,{\scriptsize(-99.9)}}][c]{76.0} & \cellcolor[HTML]{FFFFFF}76.0\,{\scriptsize(+\phantom{0}0.0)} \\
    \IfFileExists{logos/grok.png}{\raisebox{-0.2ex}{\includegraphics[height=2ex]{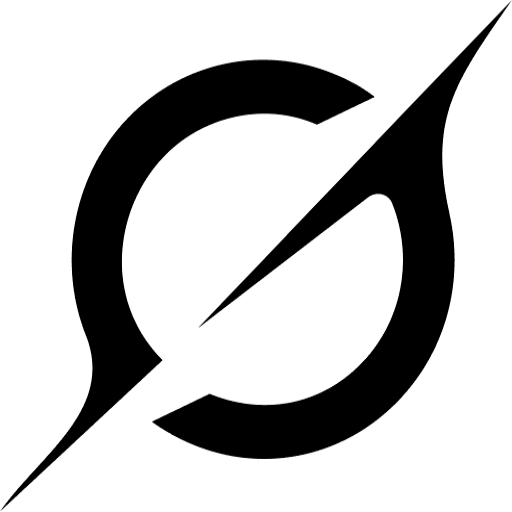}}\hspace{3pt}}{}Grok 4.20 & \cellcolor[HTML]{FFFFFF}\makebox[\widthof{99.0\,{\scriptsize(-99.9)}}][c]{97.5} & \cellcolor[HTML]{FCD6CC}79.5\,{\scriptsize(-18.5)} & \cellcolor[HTML]{FFFFFF}\makebox[\widthof{99.0\,{\scriptsize(-99.9)}}][c]{75.0} & \cellcolor[HTML]{FEF0CF}67.0\,{\scriptsize(-10.7)} & \cellcolor[HTML]{FFFFFF}\makebox[\widthof{99.0\,{\scriptsize(-99.9)}}][c]{50.0} & \cellcolor[HTML]{E39A9A}33.0\,{\scriptsize(-34.0)} & \cellcolor[HTML]{FFFFFF}\makebox[\widthof{99.0\,{\scriptsize(-99.9)}}][c]{84.0} & \cellcolor[HTML]{CC8080}\phantom{0}0.0\,{\scriptsize(-100.)} \\
    \IfFileExists{logos/moonshot.png}{\raisebox{-0.2ex}{\includegraphics[height=2ex]{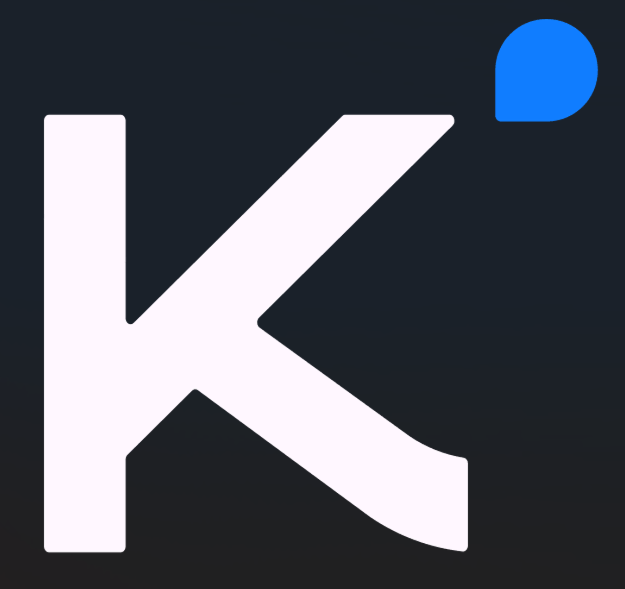}}\hspace{3pt}}{}Kimi K2.5 & \cellcolor[HTML]{FFFFFF}\makebox[\widthof{99.0\,{\scriptsize(-99.9)}}][c]{97.0} & \cellcolor[HTML]{F9C7C7}75.5\,{\scriptsize(-22.2)} & \cellcolor[HTML]{FFFFFF}\makebox[\widthof{99.0\,{\scriptsize(-99.9)}}][c]{72.0} & \cellcolor[HTML]{FEF4D8}66.0\,{\scriptsize(-\phantom{0}8.3)} & \cellcolor[HTML]{FFFFFF}\makebox[\widthof{99.0\,{\scriptsize(-99.9)}}][c]{44.0} & \cellcolor[HTML]{E69D9D}30.0\,{\scriptsize(-31.8)} & \cellcolor[HTML]{FFFFFF}\makebox[\widthof{99.0\,{\scriptsize(-99.9)}}][c]{82.0} & \cellcolor[HTML]{FEE6D1}70.0\,{\scriptsize(-14.6)} \\
    \IfFileExists{logos/moonshot.png}{\raisebox{-0.2ex}{\includegraphics[height=2ex]{logos/moonshot}}\hspace{3pt}}{}Kimi K2.6 & \cellcolor[HTML]{FFFFFF}\makebox[\widthof{99.0\,{\scriptsize(-99.9)}}][c]{96.5} & \cellcolor[HTML]{FCD9CE}79.5\,{\scriptsize(-17.6)} & \cellcolor[HTML]{FFFFFF}\makebox[\widthof{99.0\,{\scriptsize(-99.9)}}][c]{75.0} & \cellcolor[HTML]{FEF3D3}68.0\,{\scriptsize(-\phantom{0}9.3)} & \cellcolor[HTML]{FFFFFF}\makebox[\widthof{99.0\,{\scriptsize(-99.9)}}][c]{55.0} & \cellcolor[HTML]{FEF8EB}52.0\,{\scriptsize(-\phantom{0}5.5)} & \cellcolor[HTML]{FFFFFF}\makebox[\widthof{99.0\,{\scriptsize(-99.9)}}][c]{86.0} & \cellcolor[HTML]{FDDFCF}72.0\,{\scriptsize(-16.3)} \\
    \IfFileExists{logos/mistral.png}{\raisebox{-0.2ex}{\includegraphics[height=2ex]{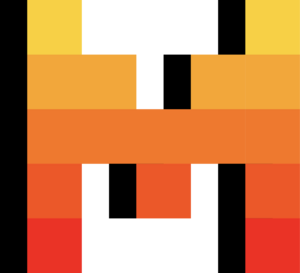}}\hspace{3pt}}{}Mistral Large 3 & \cellcolor[HTML]{FFFFFF}\makebox[\widthof{99.0\,{\scriptsize(-99.9)}}][c]{95.5} & \cellcolor[HTML]{F8C3C3}73.5\,{\scriptsize(-23.0)} & \cellcolor[HTML]{FFFFFF}\makebox[\widthof{99.0\,{\scriptsize(-99.9)}}][c]{61.0} & \cellcolor[HTML]{FEF4D9}56.0\,{\scriptsize(-\phantom{0}8.2)} & \cellcolor[HTML]{FFFFFF}\makebox[\widthof{99.0\,{\scriptsize(-99.9)}}][c]{17.0} & \cellcolor[HTML]{CC8080}\phantom{0}5.0\,{\scriptsize(-70.6)} & \cellcolor[HTML]{FFFFFF}\makebox[\widthof{99.0\,{\scriptsize(-99.9)}}][c]{56.0} & \cellcolor[HTML]{CC8080}\phantom{0}0.0\,{\scriptsize(-100.)} \\
    \bottomrule
  \end{tabular}
\end{table*}

\section{Experiments}

\begin{figure*}[t]
    \centering
    \includegraphics[width=\linewidth]{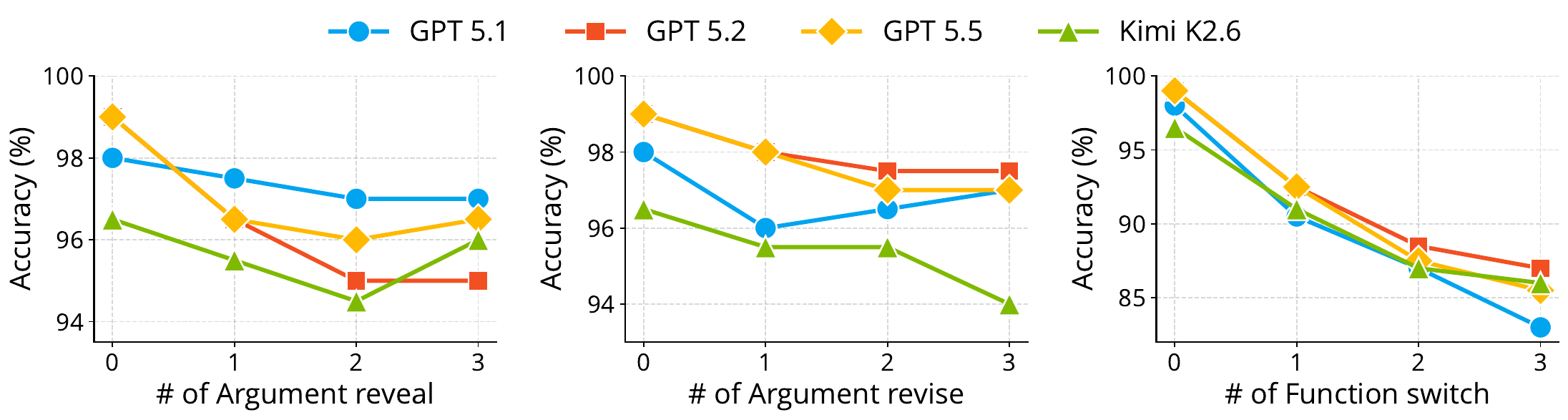}
    \vspace{-0.25in}
    \caption{\textbf{Scaling intent transitions monotonically degrades accuracy.} Accuracy of GPT 5.1, 5.2, 5.5, and Kimi K2.6 on GSM8K, where the number of intent transitions of each type (i.e., argument reveal, argument revision, and function switch) is varied independently per conversation.}
    \label{fig:scaling}
    \vspace{-0.1in}
\end{figure*}

We provide an empirical evaluation of  by investigating the following questions:
\begin{itemize}[leftmargin=*,topsep=1pt,itemsep=.5pt]
\item Does LLM performance on single-turn benchmarks transfer to the multi-turn evolving intent scenarios? (\Cref{tab:main}) 
\item How do the type, count, composition, and order of intent transitions affect performance? (\Cref{fig:scaling}, \Cref{tab:six_scenarios_ablation}, and \Cref{fig:order})
\item To what extent can practical mitigation strategies recover performance? (\Cref{fig:memory})
\item How does the difficulty of the underlying single-turn task affect performance? (\Cref{fig:difficulty})
\end{itemize}

\paragraph{Source single-turn datasets.}
We select the subset of four verifiable single-turn benchmarks: GSM8K (math, 200 samples; \citealp{cobbe2021training}), BIRD-SQL (text-to-SQL, 100 samples; \citealp{li2023can}), BrowseComp+ (agentic search, 100 samples; \citealp{chen2026browsecomp}), and SWE-Bench Verified (software engineering, 50 samples; \citealp{jimenez2024swe}). For each dataset, we use domain-specific prompts to (i) extract intents and (ii) generate counterfactual arguments and predecessor functions, all run with GPT 5.1 \citep{openai2025gpt5}. All evaluations report accuracy under each dataset's native verifier; see \Cref{app:dataset_details} for more details.

For SWE-Bench Verified, each policy is wrapped in a mini-SWE-agent v2 scaffold \citep{yang2024sweagent}, extended to multi-turn by intercepting the agent's submission and injecting the next scripted user turn until the script is exhausted, with a per-turn tool-call budget of 100 (We raised the per-turn budget to 200 for Kimi K2.6 and DeepSeek V3.2, which more readily hit the limit.). For BrowseComp+, we cap per-turn search calls at 50. 

\paragraph{Evolving intent scenario.}
For the main experiment, we evaluate the evolving intent scenario built from three transition types; reveal, revision, and switch. We combine these transition types such that each occurs twice within a conversation, resulting in up to 7 turns, including the initial turn. 
While the dynamics naturally extend to longer conversations, we use 6 transitions for tractability. 

\paragraph{LLM agents.}
We evaluate closed source LLMs, GPT 5.1, 5.2 and 5.5 \citep{openai2025gpt5}, Gemini 3.1 Pro \citep{google2026gemini31pro}, and Grok 4.20 \citep{xai2026grok420}, together with the open weight LLMs Kimi K2.5 and K2.6 \citep{moonshot2026kimik26}, Mistral Large 3 \citep{mistral2026ministral3}, and DeepSeek V3.2 \citep{deepseekai2025deepseekv32}. All models are run at default reasoning for each turn (medium, where applicable) with greedy decoding where supported.

\subsection{Main Simulation Results} 

We present our main result by simulating multi-turn conversations across four datasets under six scenarios. Here, we mainly report the degradation induced by evolving intent, measured by comparing the multi-turn performance against the single-turn source problem. As shown in \Cref{tab:main}, in most cases, strong single-turn performance does not carry over to the multi-turn conversation setting. In particular, even models that excel in the single-turn setting degrade substantially once intent evolves during the conversation, with relative drops reaching up to roughly 30\% from their single-turn accuracy (e.g., DeepSeek V3.2 on BIRD). We believe the accuracy drop is severe because the model requires to accumulate information across user turns, but also to selectively attend to updated intents while discarding prior information, making it harder to maintain a faithful internal belief.

Furthermore, SWE-Bench reveals an extreme degradation regime: several agents, such as GPT 5.1, easily exhaust the 100 tool-call budget, lingering in extended thinking and timing out on evolving setting. 
More broadly, degradation is particularly large in Search and SWE despite the increased total tool-call opportunities in the multi-turn setting, i.e., the number of turns multiplied by the per-turn tool budget. 
A closer look in SWE domain reveals that the model spends most budget on search rather than execution: fewer than 4 of 100 calls per turn are execution-related (e.g., \texttt{pytest}, \texttt{python}, \texttt{apply\_edits()}), while the rest are devoted to exploration (\texttt{sed}, \texttt{grep}, \texttt{find/rg}, \texttt{ls}).
This suggests that more tool access does not necessarily improve adaptation to evolving intent, since the accumulated interaction context and tool traces can themselves become distractors. 

\begin{table*}[t]
  \centering
  \small
  \caption{\textbf{Models struggle more under increasingly complex evolving-intent compositions.} Ablation of intent transition composition on GPT 5.1 and 5.5 across four datasets, from single-turn interactions to multi-transition compositions involving argument reveal, argument revise, and function switch dynamics. Each transition type is applied twice, with compositions ranging from two intent transitions (three turns) to six intent transitions (seven turns, including the initial turn). Background color indicates the degradation from each column's single-turn accuracy (saturating at 50\% relative drop).}
  \vspace{-0.1in}
  \label{tab:six_scenarios_ablation}
  \setlength{\tabcolsep}{9pt}
  \begin{tabular}{lcccccccc}
    \toprule
     & \multicolumn{2}{c}{\textbf{GSM8K}} & \multicolumn{2}{c}{\textbf{BIRD-SQL}} & \multicolumn{2}{c}{\textbf{BrowseComp+}} & \multicolumn{2}{c}{\textbf{SWE-Bench Verif.}} \\
    \cmidrule(lr){2-3} \cmidrule(lr){4-5} \cmidrule(lr){6-7} \cmidrule(lr){8-9}
    Scenario & \IfFileExists{logos/openai.png}{\raisebox{-0.2ex}{\includegraphics[height=2ex]{logos/openai}}\hspace{3pt}}{}5.1 & \IfFileExists{logos/openai.png}{\raisebox{-0.2ex}{\includegraphics[height=2ex]{logos/openai}}\hspace{3pt}}{}5.5 & \IfFileExists{logos/openai.png}{\raisebox{-0.2ex}{\includegraphics[height=2ex]{logos/openai}}\hspace{3pt}}{}5.1 & \IfFileExists{logos/openai.png}{\raisebox{-0.2ex}{\includegraphics[height=2ex]{logos/openai}}\hspace{3pt}}{}5.5 & \IfFileExists{logos/openai.png}{\raisebox{-0.2ex}{\includegraphics[height=2ex]{logos/openai}}\hspace{3pt}}{}5.1 & \IfFileExists{logos/openai.png}{\raisebox{-0.2ex}{\includegraphics[height=2ex]{logos/openai}}\hspace{3pt}}{}5.5 & \IfFileExists{logos/openai.png}{\raisebox{-0.2ex}{\includegraphics[height=2ex]{logos/openai}}\hspace{3pt}}{}5.1 & \IfFileExists{logos/openai.png}{\raisebox{-0.2ex}{\includegraphics[height=2ex]{logos/openai}}\hspace{3pt}}{}5.5 \\
    \midrule
    Single & \cellcolor[HTML]{FFFFFF}98.0 & \cellcolor[HTML]{FFFFFF}99.0 & \cellcolor[HTML]{FFFFFF}72.0 & \cellcolor[HTML]{FFFFFF}80.0 & \cellcolor[HTML]{FFFFFF}49.0 & \cellcolor[HTML]{FFFFFF}65.0 & \cellcolor[HTML]{FFFFFF}72.0 & \cellcolor[HTML]{FFFFFF}86.0 \\
    Argument reveal & \cellcolor[HTML]{FFFEFB}97.0 & \cellcolor[HTML]{FFFBF4}96.0 & \cellcolor[HTML]{FFFFFF}72.0 & \cellcolor[HTML]{FEF9EE}76.0 & \cellcolor[HTML]{FEF4D9}45.0 & \cellcolor[HTML]{EAA5A5}46.0 & \cellcolor[HTML]{E7A0A0}50.0 & \cellcolor[HTML]{FEF6E1}80.0 \\
    Argument revise & \cellcolor[HTML]{FFFDF9}96.5 & \cellcolor[HTML]{FFFDF8}97.0 & \cellcolor[HTML]{FFFFFF}72.0 & \cellcolor[HTML]{FEFAF2}77.0 & \cellcolor[HTML]{FEFAF1}47.0 & \cellcolor[HTML]{FEF7E6}61.0 & \cellcolor[HTML]{FEE7D0}62.0 & \cellcolor[HTML]{FFFFFF}90.0 \\
    Function switch & \cellcolor[HTML]{FEEECF}87.0 & \cellcolor[HTML]{FEEDCF}87.5 & \cellcolor[HTML]{FCD7CD}59.0 & \cellcolor[HTML]{FCD4CC}65.0 & \cellcolor[HTML]{FEF4D9}45.0 & \cellcolor[HTML]{FEF9EF}62.0 & \cellcolor[HTML]{CC8080}0.0 & \cellcolor[HTML]{FEF9EF}82.0 \\
    Revise + Switch & \cellcolor[HTML]{FDDFCF}82.0 & \cellcolor[HTML]{FEE6D1}84.5 & \cellcolor[HTML]{FEF2D1}65.0 & \cellcolor[HTML]{FEE5D1}68.0 & \cellcolor[HTML]{E59C9C}33.0 & \cellcolor[HTML]{FCDCCF}54.0 & \cellcolor[HTML]{CC8080}0.0 & \cellcolor[HTML]{FEF6E1}80.0 \\
    Reveal + Revise + Switch & \cellcolor[HTML]{FDDFCF}82.0 & \cellcolor[HTML]{FCD5CC}80.5 & \cellcolor[HTML]{FEF4D8}66.0 & \cellcolor[HTML]{FEEECF}71.0 & \cellcolor[HTML]{E7A0A0}34.0 & \cellcolor[HTML]{FEEBCF}57.0 & \cellcolor[HTML]{CC8080}0.0 & \cellcolor[HTML]{FEF6E1}80.0 \\
    \bottomrule
  \end{tabular}
\end{table*}

\subsection{Ablation Study}

We now conduct an ablation study of intent transition dynamics along multiple dimensions, including transition type, transition count, transition composition, and simulation order.

\paragraph{Scaling across transition types.}
We analyze how performance changes as we scale the number of intent transitions of each type, evaluating GPT 5.1, 5.2, 5.5, and Kimi K2.6 on GSM8K. As shown in \Cref{fig:scaling}, increasing the count of any transition type leads to a gradual drop in accuracy across all models. Notably, function switch causes a steeper decline than argument reveal and revision, which we again attribute to the larger belief state update it demands (i.e., the agent has to disregard more of the previously accumulated context rather than just absorb new information). While argument reveal is naturally bounded by the number of arguments, our newly defined argument revision and function switch can be scaled indefinitely, enabling the simulation of long-horizon evolving-intent interactions.

\paragraph{Composition of intent transitions.}
We next study how well LLMs can track user intent and act correctly under dynamically composed intent transitions. To this end, we progressively increase the compositional complexity of evolving-intent interactions by combining different transition types and examining how performance changes. Specifically, we evaluate GPT 5.1 and 5.5 across four domains, where each transition type is applied twice, yielding scenarios ranging from isolated transitions to fully composed interactions with six intent transitions. As shown in \Cref{tab:six_scenarios_ablation}, performance generally degrades as more diverse intent transitions are composed, suggesting that models increasingly struggle to maintain accurate intent tracking under more complex conversational dynamics. In particular, function switch consistently emerges as the most challenging transition type, and this degradation becomes more pronounced when combined with other transitions. Since real-world user interactions often require identifying the current intent from context scattered across prior turns, these results highlight the need for collaborative agents that can more selectively extract relevant context and update their belief state accordingly.

\subsection{Additional Analysis and Discussion}

\begin{figure}[t]
    \centering
    \includegraphics[width=\linewidth]{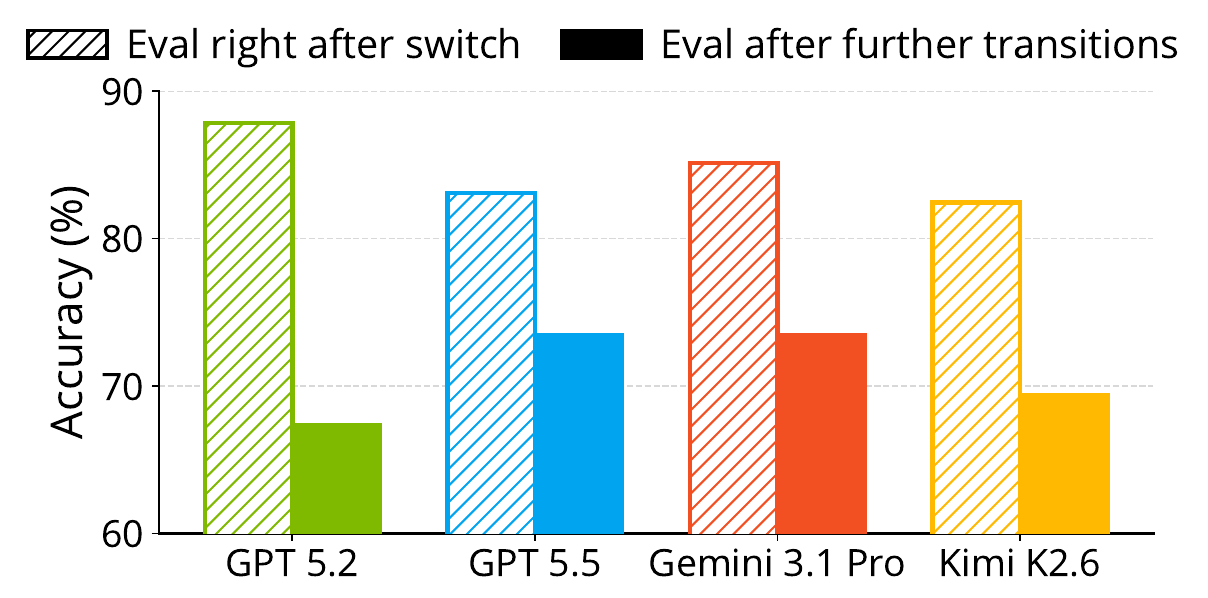}
    \vspace{-0.15in}
    \caption{\textbf{LLMs gradually shift away from earlier context after function switches.} Analysis on GSM8K across four models under evolving-intent interactions, comparing cases evaluated immediately after a function switch with those evaluated after subsequent reveal or revision transitions. Performance drops after intervening transitions, suggesting that agents struggle to jointly integrate pre-switch context with post-switch updates.}
    \label{fig:order}
    \vspace{-0.05in}
\end{figure}

\paragraph{Context carryover after function switches.}
The composition analysis above shows that function switch is particularly challenging, especially when combined with other intent transitions. We therefore take a closer look at how LLMs use conversational context around function switches. Specifically, we ask whether agents can still use the relevant context introduced before the switch while incorporating subsequent intent updates. To this end, we analyze GSM8K simulations with six intent transitions, and compare cases evaluated immediately after a function switch with cases evaluated after additional reveal or revision transitions following the switch. As shown in \Cref{fig:order}, although the two groups have similar source accuracy, models perform substantially worse when additional transitions intervene after the function switch. This suggests that agents can often recover the switched intent immediately after the switch, but struggle to jointly integrate the context before the switch with the updates introduced afterwards. 

\begin{figure}[t]
    \centering
    \includegraphics[width=\linewidth]{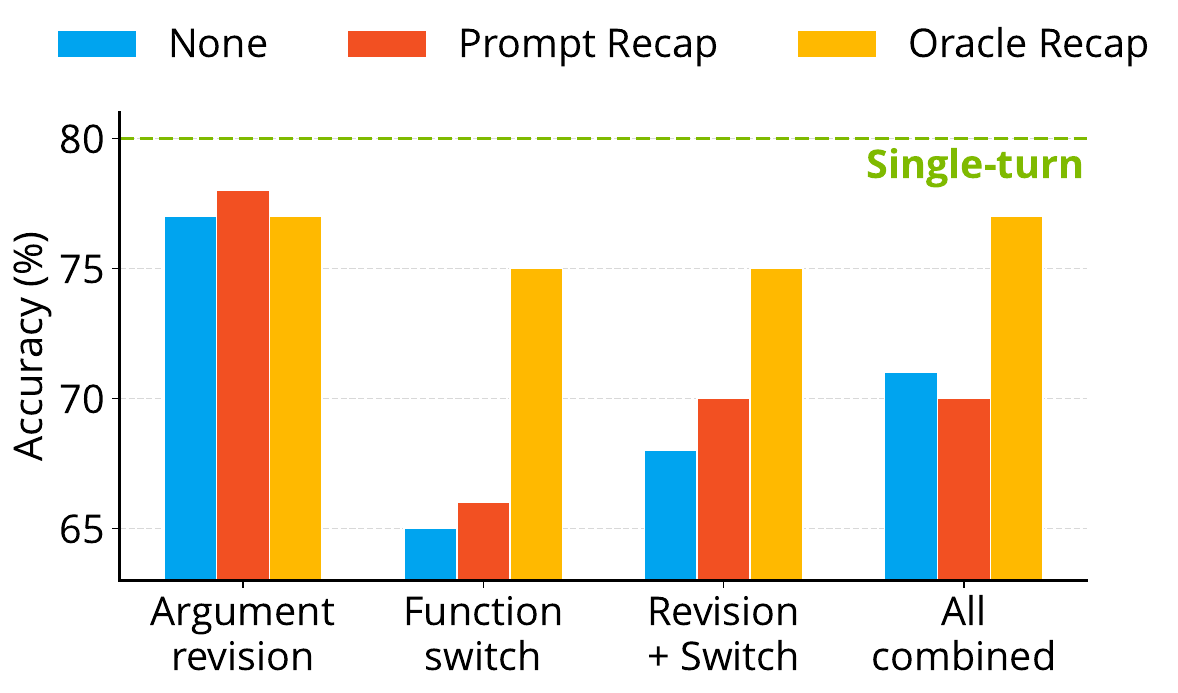}
    \vspace{-0.25in}
    \caption{\textbf{Memory mechanisms help but do not close the gap.} Accuracy of GPT 5.5 on BIRD-SQL across the evolving-intent scenarios, comparing the default agent against two memory mechanisms: prompt recap, which prepends a reminder at every turn, and oracle recap, which re-states the full revealed intent at each turn. The dashed line indicates single-turn accuracy.}
    \label{fig:memory}
\end{figure}

\paragraph{Effect of memory system.}
A natural follow-up question is whether the degradation under evolving intent can be mitigated by augmenting the agent with a simple memory mechanism, and we explore two such designs. The first, prompt recap, prepends a reminder at every turn instructing the agent to carefully revisit the prior conversation before answering. The second, oracle recap, re-states the full oracle intent revealed up to that point, i.e., the components $f_t$ and $\mathcal{C}_t^{\mathtt{rev}}$ of $\mathcal{I}_t$ (examples in \Cref{app:recap_examples}). As shown in \Cref{fig:memory}, on GPT 5.5 with BIRD-SQL, both designs yield non-trivial gains, and oracle recap in particular brings substantial improvements (e.g., from 65\% to 75\% under function switch). Yet, even with oracle recap, all scenarios still fall short of the single-turn accuracy of 80\%. We therefore believe that it is an interesting future direction to develop systems that actively recap what the user currently wants \citep{zhou2026tom}, and complementarily, reduce attention to context unrelated to the current task.

\begin{figure}[t]
    \centering
    \includegraphics[width=\linewidth]{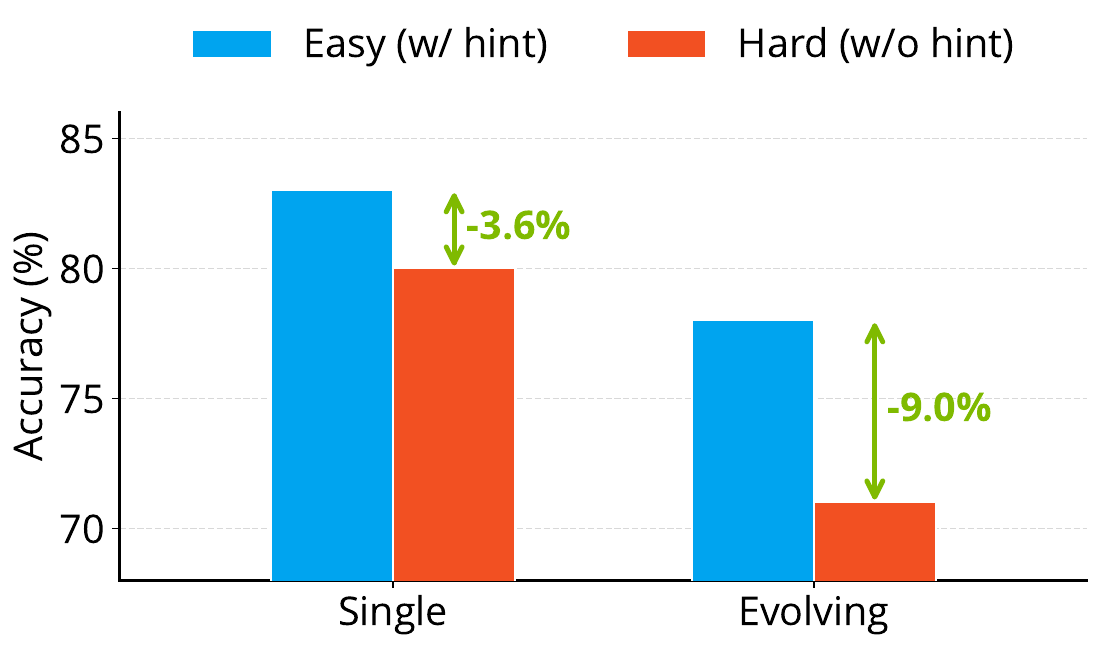}
    \vspace{-0.25in}
    \caption{\textbf{Multi-turn interaction compounds problem difficulty.} Accuracy of GPT 5.5 on BIRD-SQL, where the same queries are simulated under single-turn and evolving intent scenario. Difficulty is controlled by providing or withholding the external-knowledge hint on otherwise identical questions. Values next to the arrow indicate the relative performance drop.}
    \label{fig:difficulty}
    \vspace{0.1in}
\end{figure}

\paragraph{Effect of source data difficulty.} To understand how the difficulty of the source data affects multi-turn conversation performance, we control problem difficulty on BIRD-SQL. Specifically, BIRD-SQL requires domain expertise to translate text into SQL, and the dataset optionally provides this expertise as a hint, allowing us to vary the difficulty of the same source problem. As shown in \Cref{fig:difficulty}, with GPT 5.5, increasing difficulty hurts the multi-turn setting more than the single-turn setting (i.e., the relative drop from easy to hard is 3.6\% in single-turn but 9.0\% in multi-turn). This implies that LLMs degrade more under multi-turn dynamics when the source data is more difficult. In practice, this suggests that LLMs may struggle even further when users delegate more challenging problems with evolving intent.

\begin{table}[t]
    \caption{
        \textbf{Turn-wise intent-tracking performance} Accuracy of GPT 5.1 on GSM8K conversations with evolving user intent. At the end of each turn, agent is prompted to predict the user's current intent. A separate judge evaluates each prediction against the ground-truth intent and outputs score between 0 and 1 (we report \%).}
    \vspace{-0.1in}
    \label{tab:intent-tracking}
    \centering
    \small
    \begin{tabular}{lcc}
        \toprule
        \textbf{Transition Type} 
        & \textbf{1 Occurrence} 
        & \textbf{2 Occurrences} \\
        \midrule
        Argument Reveal & 99.0 & 98.0 \\
        Argument Change & 98.0 & 96.0 \\
        Function Switch & 89.0 & 82.0 \\
        \bottomrule
    \end{tabular}
    \vspace{-0.1in}
\end{table}

\paragraph{Turn-wise intent tracking analysis.}
Figure \ref{fig:memory} separates two sources of error in conversations with evolving user intent. 
Prompt recap evaluates whether the model can track the current intent from the conversation history, whereas oracle recap provides the correct intent and evaluates whether the model can act on it under conflicting context. 

To isolate failures in tracking evolving user intent, we conduct a turn-wise intent-tracking analysis on GSM8K. At the end of each turn, we prompt the test model to predict the user’s current intent from the conversation history. A separate judge compares the prediction against the ground-truth intent using a binary correctness score. As shown in Table~\ref{tab:intent-tracking}, intent tracking remains nearly perfect for argument reveals and changes, but degrades substantially for function switches, particularly when multiple switches occur.

\section{Conclusion}

In this paper, we study whether LLM agents can faithfully track and act on the user's intent as it evolves throughout a conversation. To this end, we propose a framework that converts any verifiable single-turn benchmark into a multi-turn environment by anchoring the source intent at the final turn and backward-synthesizing preceding turns along three controllable transitions, i.e., argument reveal, revision, and function switch, while preserving the dataset's original verifier. Across multiple domains, both frontier and open-source LLMs degrade substantially once the user's intent becomes dynamic, revealing a fundamental gap invisible to static, single-turn evaluation yet critical for future collaborative agents.

\section*{Limitations and Future Directions}

Our framework focuses on intent evolution, but does not model finer grained variation in user behavior, such as persona, communication style, typos, or grammatical errors \cite{sun2025training,naous2026flipping}. As a result, the rendered utterances may be more stylistically uniform (compared to diverse users). A natural extension is to make the pipeline persona conditioned, so that counterfactual arguments, predecessor functions, and rendered utterances reflect diverse user backgrounds and communication patterns \citep{liang2026learning}.

We also assume that each user turn contains a single intent transition, which keeps the dynamics controllable and the final verifier well defined. In practice, users may revise one part of the request while switching to another task, or ask multiple related questions in the same turn. Extending the framework to turns with multiple intents and multiple verifiable targets would broaden the range of conversational behaviors captured by the benchmark \citep{hou2026compound}.

Furthermore, our verifier is exact only at the final turn, since each conversation is constructed to converge to a source single turn problem. We view this as a deliberate tradeoff for scalability and automatic evaluation: the intermediate trajectories are still controlled by the scheduled intent transitions, and their effect is reflected in the final outcome. Designing intermediate verifiers that preserve this scalability, while allowing more flexible user intent updates, is an interesting direction for future work.

\small
\bibliography{reference}

% =====================================================
%                       Appendix
% =====================================================
\clearpage
\onecolumn
\appendix
\normalsize

\section{User-Agent Interaction}
\label{app:pomdp}

\begin{figure*}[ht]
\centering
\begin{tikzpicture}[
    >=Stealth,
    every node/.style={font=\small},
    box/.style={draw, rounded corners=4pt, minimum width=2.6cm, minimum height=1.0cm, align=center, thick},
    arr/.style={->, thick, >=Stealth},
]

\node[box, fill=gray!10] (history) {$h_{t-1}$\\[2pt]{\small Interaction History}};

\node[box, fill=blue!10, right=2.0cm of history] (user) {$\mathcal{U}$\\[2pt]{\small User Text Renderer}};

\node[box, fill=orange!10, right=2.6cm of user] (model) {$\mathcal{M}$\\[2pt]{\small LLM Agent}};

\node[box, fill=red!10, draw=red!60, minimum width=4.5cm, minimum height=1.0cm,
      above=1.2cm of user] (belief) {%
  $\mathcal{I}_t = (f_t,\, \mathcal{C}_t,\, \mathcal{C}^{\mathrm{rev}}_t,y)$\\[2pt]
  {\small User Intent State}};

\node[box, fill=green!10, above=1.2cm of model] (reward) {$R_t$\\[2pt]{\small Reward}};

\draw[arr, color=red!60] (belief) -- node[right, font=\footnotesize] {$\Delta\mathcal{I}_t$} (user);

\draw[arr, color=gray!60] (history) -- (user);

\draw[arr, color=gray!60] ([xshift=0.4cm]history.south) -- ++(0, -0.35) -| (model.south west);

\draw[arr, color=blue!70] (user) -- node[above, font=\footnotesize] {$u_t$} (model);

\draw[arr, color=orange!70] (model) -- node[right, font=\footnotesize] {$a_t$} (reward);

\draw[arr, color=red!50] (belief) -- node[above, font=\footnotesize] {$y_t$} (reward);

\end{tikzpicture}
\caption{The user-LLM agent collaboration loop as a POMDP. At user turn $t$, the user text renderer $\mathcal{U}$ observes the latent intent state $\mathcal{I}_t$ and the interaction history $h_{t-1}$, and renders a user utterance $u_t$. In our implementation, this utterance is conditioned on the updated function and the newly revealed or revised conditions. The LLM agent $\mathcal{M}$ observes only $o_t=(u_t,h_{t-1})$ and must infer the user's latent intent from the conversation. The reward $R_t$ measures alignment between the agent action $a_t$ and the latent answer $y_t$, but our benchmark only evaluates the final-turn reward so that each trajectory remains automatically verifiable against the original single-turn answer.}
\label{fig:pomdp}
\end{figure*}
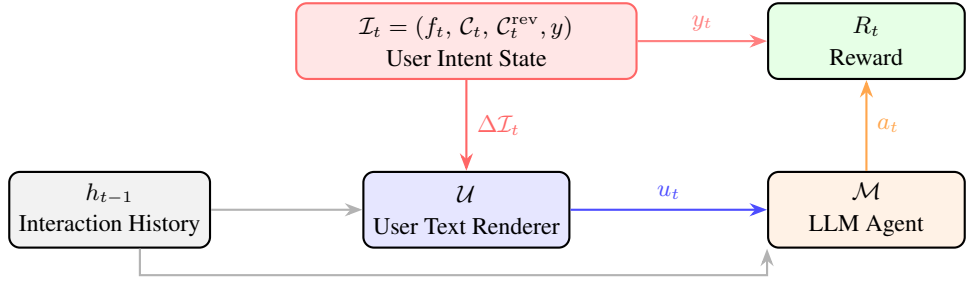

We frame user-agent collaboration as a Partially Observable Markov Decision Process (POMDP) grounded in the user's latent intent (see \Cref{fig:pomdp}). The user has a latent intent state $\mathcal{I}_t$ at each turn, which specifies the task they want to solve and the constraints revealed so far, while the agent $\mathcal{M}$ only partially observes this intent through user utterances. We denote by $\Delta\mathcal{I}_t$ the turn-level intent update, such as a newly revealed argument, a revised argument, or a function switch, that determines what information should be expressed at turn $t$. At each turn $t$, the user text renderer $\mathcal{U}$ produces a user utterance $u_t \sim \mathcal{U}(\cdot \mid \Delta\mathcal{I}_t, h_{t-1})$, where $h_{t-1}$ is the interaction history before turn $t$. In this paper, we mainly implement $\mathcal{U}$ as a rule-based text renderer conditioned on $\Delta\mathcal{I}_t$, while also providing an LLM-based renderer that can incorporate the interaction history (see \Cref{app:sim:rendering} for more details). The agent $\mathcal{M}$ then observes $o_t = (u_t, h_{t-1})$ and acts via $a_t \sim \mathcal{M}(\cdot \mid o_t)$. The history updates to $h_t = (h_{t-1}, u_t, a_t)$, and the user's latent intent transitions to $\mathcal{I}_{t+1}$ for the next turn. The agent receives reward $R_t = R(a_t, y_t)$, where $y_t$ is the target answer induced by the latent intent; in our benchmark, this reward is evaluated only at the final turn.

\section{Experimental Setup Details}
\label{app:dataset_details}

In this section, we discuss the models and datasets used in the paper. 

\subsection{Model Details}

We evaluate several closed-source LLMs, including GPT 5.1, GPT-5.2, and GPT 5.5~\citep{openai2025gpt5}, Gemini 3.1 Pro~\citep{google2026gemini31pro}, and Grok 4.20~\citep{xai2026grok420}. We also have used GPT 5.4, GPT 5.4 mini and GPT 5.4 nano in \Cref{app:more_exp} for model scaling experiment. For these closed-source models, the full parameter counts are not publicly available. For open-source models, we evaluate Kimi K2.5 and Kimi K2.6 \citep{moonshot2026kimik26}, both 1T-parameter MoE models with 32B activated parameters, and Mistral Large 3 \citep{mistral2026ministral3}, a 675B-parameter MoE model with 39B activated parameters.

\subsection{Dataset Details}

We selected four domains, including math, text-to-SQL, agentic search, and agentic software engineering, which are mainly considered single-turn benchmarks.

\paragraph{GSM8K.}
GSM8K~\citep{cobbe2021training} consists of grade-school math word problems with a single numerical answer. We extract the target function as the quantity to compute and extract the arguments from the numerical and relational premises in the problem statement. For verification, the final turn is anchored to the original GSM8K problem, so the agent's final answer can be checked against the original numerical answer after standard answer normalization.

\paragraph{BIRD-SQL.}
BIRD-SQL~\citep{li2023can} is a text-to-SQL benchmark where each instance pairs a natural-language question with a database, evidence, and a gold SQL query. For SQL tasks, we define the intent structure directly from the gold SQL. The \emph{function} corresponds to the output shape and query operation, including clauses such as \texttt{SELECT}, \texttt{GROUP BY}, \texttt{ORDER BY}, \texttt{JOIN}, and \texttt{LIMIT}; these clauses determine what to return and how to aggregate, join, sort, or truncate the result. The \emph{arguments} correspond to filtering constraints in \texttt{WHERE} and \texttt{HAVING}; each predicate is treated as one user-provided argument that specifies which rows qualify.

Based on this decomposition, we implement a SQL-to-text renderer that converts each gold SQL into a structured intent description with a function sentence and a list of argument sentences. This representation allows us to generate counterfactual arguments by changing \texttt{WHERE}/\texttt{HAVING} predicates and predecessor functions by changing the functional SQL clauses, while keeping the resulting queries executable. For evaluation, the final turn is anchored to the original SQL task, and the agent's final query is scored using execution-based verification.

\paragraph{BrowseComp+.}
BrowseComp+~\citep{chen2026browsecomp} is a benchmark of complex multi-constraint research queries paired with curated gold and distractor evidence documents, designed to evaluate deep web-search and retrieval-augmented reasoning. Each query specifies several factual constraints (e.g., a person or entity satisfying multiple temporal, geographic, and biographical conditions) and admits a single short answer. This domain naturally provides argument-rich examples, which makes it suitable for evaluating under-specified and evolving user intent. For verification, the final turn recovers the original query, and the agent's final answer is checked against the gold answer.

\paragraph{SWE-Bench Verified.}
SWE-Bench Verified~\citep{jimenez2024swe} is the human-verified subset of SWE-Bench, consisting of real GitHub issues paired with gold patches and unit tests across popular Python repositories. For each instance, we extract the \emph{function} from the high-level software change requested in the issue, and extract \emph{arguments} from concrete requirements in the cleaned problem statement, such as the observed symptom, triggering input, affected component, implementation scope, and constraints. We use only the problem statement for intent extraction, counterfactual argument generation, and predecessor function generation. The gold patch and tests are reserved for verification: we check that the final rendered turn is aligned with the original issue, and during evaluation, the agent's final patch is scored using the original SWE-Bench test-based protocol. We retain only instances that contain enough extractable arguments to support evolving-intent interactions and pass this verification pipeline.

\paragraph{Filtering and sampling.}
The retained instances are challenging because they must support evolving-intent interactions with a well-defined function, multiple arguments, counterfactual arguments, and predecessor functions, all while preserving automatic verifiability. We first filter examples through our full data processing pipeline, including intent extraction, counterfactual generation, predecessor generation, and verification. From the verified pool, we randomly sample small evaluation sets for cost-effective evaluation: 200 GSM8K, 100 BIRD-SQL, 100 BrowseComp+, and 50 SWE-Bench Verified examples. This is necessary because each 7-turn interaction requires up to seven sequential agent calls with increasingly long contexts, substantially increasing inference cost compared to single-turn evaluation.

\paragraph{Licenses and terms of use.}
We use existing public benchmarks for research evaluation and cite their original creators. For each artifact, we follow the license or terms of use specified by the original benchmark release. Our use is limited to benchmark construction and evaluation, and we do not use these artifacts for commercial deployment or human-subject data collection. 

\section{Simulating Multi-Turn Conversations}
\label{app:sim_details}

This appendix describes how we convert an abstract intent trajectory into the concrete user messages shown to an agent. Given a single-turn benchmark example, the simulator first constructs a sequence of intent states whose final state matches the original example, then schedules the corresponding intent transitions across turns, renders each turn into natural language.

\subsection{Turn Scheduling}
\label{app:sim:scheduling}
th
Given an extracted intent $(f, \mathcal{C}=\{c_1,\ldots,c_N\}, y)$ from a single-turn example, the scheduler receives three integer knobs: the number of function switches $g$, the number of argument revisions $p$, and the total number of user turns $T$. We require $T \geq 1 + g + p$, since the initial turn introduces the first intent state and each function switch or revision must occupy at least one subsequent turn. Under this schedule, the remaining $T-1-g-p$ turns act as reveal-only turns, controlling how many additional opportunities the user has to disclose arguments without changing or correcting the intent.

The scheduler outputs an ordered list of \texttt{TurnSlot}s $(S_1,\ldots,S_T)$. Each slot $S_t$ is a structured representation of the turn-level intent update $\Delta\mathcal{I}_t$: a function event changes the current function and its argument set, a revision event changes the value of an argument, and an argument reveal adds an argument to $\mathcal{C}^{\mathtt{rev}}_t$, the subset of arguments already revealed to the agent. Applying these slots sequentially induces the latent intent trajectory $\mathcal{I}_1,\ldots,\mathcal{I}_T$. No surface text is generated at this stage; surface utterances are produced later by the user text renderer conditioned on the corresponding $\Delta\mathcal{I}_t$.

The key design choice is to schedule intent transitions before rendering the conversation. This prevents invalid trajectories in which a revised argument is corrected before its perturbed value has been introduced, or in which an argument is revealed under a function to which it does not belong. Concretely, the scheduler first reserves slots for function switches and argument revisions, and then fills the remaining available slots with argument reveals while respecting their validity constraints. Thus, $g$ and $p$ determine the number of intent-changing turns, while $T-1-g-p$ determines the number of pure reveal turns. This procedure yields a valid schedule whenever the selected perturbations are available and the turn budget satisfies the feasibility constraint.

\begin{algorithm}[t]
\caption{Turn scheduling.}
\label{alg:scheduler}
\begin{algorithmic}[1]
\Require Anchor intent $(f^*, \mathcal{C}^*, y^*)$, perturbation pools, knobs $(g,p,T)$.
\Ensure Ordered list of \texttt{TurnSlot}s $(S_1,\ldots,S_T)$.
\State Select a chain of $g$ predecessor functions and $p$ counterfactual arguments from the available perturbation pools.
\State Initialize $S_1$ with the first active function and at least one argument reveal.
\State Schedule function-switch and argument-revision events over turns $2,\ldots,T$.
\State For each argument revision, set its correction turn as a deadline, so that the corresponding counterfactual value is revealed strictly before the revision occurs.
\State Fill the remaining available positions with argument reveals, deferring reveals when possible but never past their deadlines or under the wrong active function.
\State Within each turn, order events as function event $\succ$ revision event $\succ$ argument reveal.
\State Return the resulting sequence of \texttt{TurnSlot}s.
\end{algorithmic}
\end{algorithm}

\subsection{Text Rendering}
\label{app:sim:rendering}

After scheduling, each \texttt{TurnSlot} contains a structured intent update, such as an updated function, an argument correction, or newly revealed arguments. The rendering layer converts this structured representation into a single user utterance. We support two interchangeable rendering backends.

\paragraph{Rule-based rendering.}
The default renderer is rule-based. It concatenates the structured content with short prefix phrases sampled from domain-specific banks. The prefix banks are organized by transition type, so that function switches, argument revisions, and argument reveals receive different surface forms. For example, revision turns may start with phrases such as ``Wait, I need to correct that,'' while reveal turns may start with phrases such as ``One more thing'' or ``I forgot to mention.'' SQL examples use filter-oriented phrasing, while math and instruction-following examples use more general conversational phrasing. This renderer preserves every load-bearing value exactly, and is deterministic except for prefix sampling. Therefore, it is the default backend for our main benchmark results.

\paragraph{LLM-based naturalization.}
For analyses where more natural surface forms are desirable, we also provide an LLM-based naturalizer, which we use for the BrowseComp+ experiments. The naturalizer rewrites each rule-based turn into a shorter and more conversational user message, optionally conditioned on the previous assistant response. This makes the conversation less template-like; for example, the user can briefly acknowledge the previous answer before correcting a requirement.

To prevent the naturalizer from changing the underlying intent, we validate each generated utterance against the rule-based reference. The validator extracts critical tokens, such as numbers, equations, named entities, quoted strings, and constraint values, and checks that they are preserved in the rewritten turn. If validation fails, the naturalizer retries with a stricter prompt that lists the missing tokens explicitly; if it fails again, the simulator falls back to the rule-based turn.

\begin{center}
\fcolorbox{black!30}{gray!8}{%
  \begin{minipage}[t]{0.92\linewidth}\raggedright\footnotesize
  \textbf{Rule-based turn:}
  Wait, I need to correct a few things. A regular hexagon can be divided into eight equilateral triangles. The perimeter of one of the equilateral triangles is 21 inches.
  \end{minipage}}
\end{center}

\begin{center}
\fcolorbox{black!30}{blue!4}{
  \begin{minipage}[t]{0.92\linewidth}\raggedright\footnotesize
  \textbf{Naturalized turn:}
  Sorry, ignore the last bit. I had two numbers swapped: it is actually eight equilateral triangles, and the perimeter of each one is 21 inches.
  \end{minipage}}
\end{center}

\subsection{Memory Mechanism Examples}
\label{app:recap_examples}

Here, we explain the memory mechanism examples that uses i) prompting to make LLMs to improve the attention of the previous context, or ii) explicitly retrieve the oracle memory (i.e., user intent at the current turn). Each memory methods are appended to user turns after the initial turn, since there is no prior context to recall at turn $0$. 

\paragraph{{Prompt} recap.}
The cheapest intervention: a single fixed sentence appended to every non-initial turn that asks the model to re-derive the relevant context on its own, without any oracle assistance. The exact sentence we use is
shown below.

\begin{center}
\fcolorbox{black!30}{gray!8}{
  \begin{minipage}[t]{0.92\linewidth}\raggedright\footnotesize
  \textbf{User turn with \textsc{Prompt} recap:}\\
  Hold on, instead of listing them, I want the count.\\[2pt]
  \textit{Before answering, first list all the conditions from our conversation that apply to this question, then provide your answer.}
  \end{minipage}}
\end{center}

\paragraph{Oracle recap.}
At each turn, the simulator identifies the active function, selects the arguments relevant to that function, removes arguments that have not yet been revealed, and substitutes the latest active value for any revised argument.

\begin{center}
\fcolorbox{black!30}{gray!8}{%
  \begin{minipage}[t]{0.92\linewidth}\raggedright\footnotesize
  \textbf{User turn without recap:}\\
  Hold on, instead of listing them, I want the count.
  \end{minipage}}
\end{center}

\begin{center}
\fcolorbox{black!30}{blue!4}{
  \begin{minipage}[t]{0.92\linewidth}\raggedright\footnotesize
  \textbf{User turn with \textsc{Oracle} recap:}\\
  Hold on, instead of listing them, I want the count.\\[2pt]
  \textit{Before answering, recall that the goal is: ``Among the account
  opened, how many customers are there?'' The following conditions apply: the customers should be \textbf{female}; their year of birth should be before \textbf{1950}; the district is \textbf{Sokolov}.}
  \end{minipage}}
\end{center}

\vspace{0.15in}
\section{Verification of Simulation Components}
\label{app:verification}

Our simulator relies on three LLM-generated components introduced in \Cref{sec:method:anchor,sec:method:synthesis}: $\mathtt{Extract}$, $\mathtt{Counterfact}$, and $\mathtt{Predecessor}$. Since these components are used to construct synthetic multi-turn trajectories, we verify each component before accepting a sample. The goal of verification is to preserve the final turn anchor. Earlier turns may reveal incomplete information, introduce a wrong argument to be corrected later, or ask a different predecessor task. Thus, we provide a verification stage to perform rejection sampling. For all verification, we use GPT 5.1 with reasoning medium mode.

\subsection{Verifying the Extracted Anchor Intent}
\label{app:verification:extract}

The $\mathtt{Extract}$ operator defines the anchor intent that the entire synthetic dialogue must ultimately recover. Therefore, extraction errors are especially harmful: if the extracted function or arguments do not faithfully represent the source question, the simulator is no longer constructing a multi-turn version of the original problem, but a trajectory for a different problem. We therefore verify that the extracted intent preserves both the information needed to solve the source question and the original answer.

\paragraph{Coverage.}
We first verify that the extracted function and arguments contain all information required to solve the source question. An LLM judge inspects $q$, $f^*$, and $\mathcal{C}^*$, and rejects the sample if any necessary condition is missing. This prevents cases where the extractor produces an under-specified version of the original problem.

\paragraph{Solvability.}
We then verify that the extracted intent is answer-equivalent to the source question. We render $(f^*, \mathcal{C}^*)$ into natural language and ask a reference solver to answer both the rendered intent and the original question $q$. We accept the extraction if the solver produces the same answer for the rendered intent and the original question $q$, or if the solver's answer on the rendered intent matches the original gold answer $y^*$. This check filters out extractions that are complete at the surface level but change the answer of the underlying task.

\subsection{Verifying Counterfactual Arguments}
\label{app:verification:counterfact}

The $\mathtt{Counterfact}$ operator is used to create revision turns, where the user first provides an incorrect detail and later corrects it. This dynamic is meaningful only when the counterfactual is a valid, localized wrong value for an otherwise unchanged argument: if it rewrites the argument broadly, deletes information, or introduces an incompatible condition, then the resulting turn is no longer a clean revision of the original intent. We therefore verify that each counterfactual corresponds to a minimal value substitution that can be explicitly revised back to the source argument.

For each source argument $c_i^*$, $\mathtt{Counterfact}$ returns a counterfactual argument $c_i^{\mathtt{cf}}$, the original value $v_i^*$, and the replacement value $v_i^{\mathtt{cf}}$. We use an LLM verifier to accept the counterfactual only if $c_i^{\mathtt{cf}}$ can be obtained by replacing one localized value $v_i^*$ in $c_i^*$ with $v_i^{\mathtt{cf}}$, after light normalization such as whitespace and apostrophes, while preserving the surrounding argument structure. We additionally reject edits that are effectively additive or deletive, such as replacing \emph{``48''} with \emph{``48 of her closest friends''}, and require the replacement value to stay within a bounded length ratio. These checks ensure that the resulting revision is a clean value correction rather than a broader rewrite.

\subsection{Verifying Predecessor Functions}
\label{app:verification:predecessor}

The $\mathtt{Predecessor}$ operator creates earlier tasks for task switching. A valid predecessor should be related enough to the final source task to share context, but it should not collapse into the same function. In addition, any arguments introduced only for the predecessor must not change the answer to the final anchored task. We therefore verify both the semantic validity of the predecessor function and the independence of its newly introduced arguments.

\paragraph{Function validity.}
We first check whether the predecessor function is a plausible related task. An LLM judge compares $f^{\mathtt{pre}}$ with the final function $f^*$ and rejects the sample if the two tasks do not share a reasonable scenario, entity, or domain context. The judge also rejects cases where the predecessor computes the same quantity as the final function, since such cases is not a meaningful task switch.

\paragraph{Answer preservation.}
We then verify that predecessor-only arguments do not affect the final anchored task. Let $\mathcal{C}^{\mathtt{new}} = \mathcal{C}^{\mathtt{pre}} \setminus \mathcal{C}^*$ denote the arguments introduced only for the predecessor. We ask a reference solver (i.e., GPT 5.1) to answer the final function using both the original final intent $(f^*, \mathcal{C}^*)$ and the augmented intent $(f^*, \mathcal{C}^* \cup \mathcal{C}^{\mathtt{new}})$. We accept the predecessor only if the two solver outputs match, or if the answer from the augmented intent matches the original gold answer $y^*$. This is analogous to the solvability check used for anchor extraction, but here it ensures that fabricated predecessor arguments do not alter the answer of the final source problem.

\paragraph{Cross-turn independence.}
For chains with multiple predecessor tasks, we further check that arguments fabricated for one predecessor do not affect the answer to another task in the chain. For each predecessor-only argument set $\mathcal{C}_i^{\mathtt{new}}$, we ask a reference solver whether adding it to another intent changes that intent's answer. We reject the chain if the solver output changes, since this indicates that information introduced for one turn leaks into or constrains another turn. This extends the same answer-preservation check beyond the final source task to all task-switching turns.

\section{Prompt Templates}
\label{app:prompts}
We illustrate the three LLM-driven pipeline stages: extraction, predecessor function synthesis, and counterfactual argument synthesis. For each domain, we prompt the LLM with domain-specific prompts. Here, we present the BrowseComp+ (search) prompt to provide a high-level illustration. Each prompt below is reproduced verbatim, except for the few-shot example blocks, which are abbreviated as \texttt{<few-shot examples omitted for length>}. The prompts for the other domains follow the same structure and are omitted for brevity.  In all cases, we use GPT 5.1 as the underlying LLM, and we also provide a representative output for each step.

\subsection{Extraction Prompts}
\label{app:extraction_prompt}

\begin{Verbatim}[fontsize=\scriptsize,frame=single,framesep=2mm,breaklines=true,breakanywhere=true,breaksymbolleft={},breaksymbolright={},baselinestretch=0.95]
You are given a complex research query that requires finding a specific piece of information (a name, date, fact, etc.) by satisfying multiple constraints. Decompose it into a GOAL and CONDITIONS.

Output a JSON object with this format:
{
  "goal": "The main question - what to find/identify",
  "conditions": [
    {"condition": "first constraint or identifying criterion"},
    {"condition": "second constraint or identifying criterion"},
    ...
  ]
}

Rules:
1. GOAL: The main question - WHAT to find/identify
   - Should describe what the query asks WITHOUT including the specific constraints
   - The goal MUST be self-contained: do NOT use "the described person", "the above institution", "this entity", or similar references to conditions
   - Do NOT use pronouns (she, he, it, they) without a named subject in the goal itself
   - Describe what to find abstractly (e.g., "What is the name of the individual?", "Identify the learning institution.")
   - Reason: the goal is presented FIRST in conversation, before any conditions are revealed, so it must make sense alone
2. CONDITIONS: The factual constraints needed to narrow down the answer (2-8 items)
   - Each condition is a factual constraint extracted from the query
   - Conditions contain dates, locations, attributes, relationships, and identifying details
   - Preserve ALL specific details exactly (names, dates, numbers, places)
3. Each condition should be a direct quote or close paraphrase from the query text
4. Conditions should not overlap
5. Do NOT add hints, search strategies, or reasoning - only extract information from the query
6. Do NOT add INFERRED or DERIVED conditions --- only extract facts explicitly stated in the query
   - If the query doesn't explicitly say it, do NOT include it as a condition
7. Do NOT create META-CONDITIONS that describe the query itself rather than providing constraints
   - A condition must NEVER start with "The question refers", "The query asks", "The focus is on", "This pertains to"
   - If something is part of what is being ASKED, it belongs in the GOAL, not conditions
8. Each condition MUST be self-contained and understandable on its own, without reading the other conditions
   - Do NOT use back-references like "that same person", "the aforementioned", "this institution", "the event mentioned above", "the same city", "the article discussed above"
   - If a condition refers to an entity or event described elsewhere, repeat the necessary context
   - Example: Instead of "They also published a paper in 2012", write "The individual published an article in 2012"
   - Example: Instead of "in the same city", write "in the city where there was a tower built in the 1340s"
   - Think of conditions as an UNORDERED SET of independent facts --- each one must make sense if read in isolation
9. Do NOT over-split: closely related facts about the same event or criterion should stay in one condition
10. Temporal or causal relationships between conditions should be made explicit within each condition where needed

IMPORTANT: Clearly separate WHAT to find (goal) from the IDENTIFYING CRITERIA (conditions):
- "What is the name of the person who did X and Y?" should become:
  - Goal: "What is the name of the person?"
  - Condition 1: "The person did X"
  - Condition 2: "The person did Y"

<few-shot examples omitted for length>

Now decompose this query:

Q: [[QUESTION]]
\end{Verbatim}

\begin{center}
\fcolorbox{black!30}{gray!8}{%
  \begin{minipage}[t]{0.92\linewidth}\raggedright\footnotesize
  \textbf{Input} (raw BrowseComp question):\\
  I am looking for the name of an ancient village that is part of a National
  Historic Landmark dating back to A.D.\ 1350. This landmark includes a trail
  measuring 1,100--1,500 feet. As of December 2023, it is located 34--36
  miles in aerial distance from Santa Fe Regional Airport.
  \end{minipage}}
\end{center}
\begin{center}
\fcolorbox{black!30}{blue!4}{%
  \begin{minipage}[t]{0.92\linewidth}\raggedright\footnotesize
  \textbf{Output} (extracted function and three arguments):\\[2pt]
  \textbf{Function:}~What is the name of the ancient village?\\[2pt]
  \textbf{Argument 1:}~The ancient village is part of a National Historic Landmark
  dating back to A.D.\ 1350.\\[2pt]
  \textbf{Argument 2:}~The National Historic Landmark that includes the ancient
  village contains a trail measuring 1,100--1,500 feet.\\[2pt]
  \textbf{Argument 3:}~As of December 2023, the National Historic Landmark that
  includes the ancient village is located 34--36 miles in aerial distance
  from Santa Fe Regional Airport.
  \end{minipage}}
\end{center}

\clearpage
\subsection{Synthesize Prompts for Counterfactual Argument}
\label{app:counterfact_prompt}
\begin{Verbatim}[fontsize=\scriptsize,frame=single,framesep=2mm,breaklines=true,breakanywhere=true,breaksymbolleft={},breaksymbolright={},baselinestretch=0.95]
You are given a complex search/research problem with multiple conditions. Your task is to generate a PERTURBED version of the target condition.

"Perturbed" means REPLACING one specific detail with a GENUINELY DIFFERENT value so the condition points to a DIFFERENT answer. The perturbed condition must remain plausible and relevant.

KEY PRINCIPLE: The perturbed value must create a REAL FACTUAL CONTRADICTION with the original. Someone hearing the original value after the perturbed value must recognize that information ACTUALLY CHANGED. Synonyms, near-synonyms, rephrases, and subset/superset values are all FORBIDDEN --- they do not create real contradictions.

Requirements:
1. REPLACE exactly ONE detail with a different value. This can be:
   - A year or date (e.g., 2016 -> 2017)
   - A number or quantity (e.g., "eight" -> "ten", "42 months" -> "36 months")
   - A temporal range endpoint (e.g., "between 2015 and 2020" -> "between 2015 and 2018")
   - A descriptive attribute or category (e.g., "horror thriller" -> "psychological thriller", "grocery delivery" -> "meal delivery")
   - A relative temporal phrase (e.g., "early 1990s" -> "late 1990s")
   - A specific qualifier (e.g., "three-day event" -> "two-day event", "gold medal" -> "silver medal")
2. The change should be PLAUSIBLE and CONTEXTUALLY APPROPRIATE --- it must make sense within the same domain
3. Keep the EXACT same sentence structure --- the perturbed condition must have the same length and form as the original, with only one value swapped
4. CRITICAL: Do NOT change the main subject/entity being searched for
   - If the question is about a specific person, company, or event, the perturbed condition must still be about that same entity
   - Only change a QUALIFYING DETAIL, not the core subject
5. The perturbed value should be close but different --- a nearby alternative that is realistic
6. CRITICAL: The perturbation must be a VALUE SWAP, NOT an addition or elaboration:
   - CORRECT: "road accident" -> "motorcycle accident" (swaps "road" for "motorcycle")
   - CORRECT: "2018" -> "2019" (swaps year value)
   The reason: in our evaluation framework, the original condition will later REPLACE the perturbed one. If the perturbation ADDS information, the original cannot contradict it --- the added detail becomes an orphan that persists. Only a value SWAP ensures the original directly contradicts the perturbed version.
7. CRITICAL: The perturbed condition must be obtainable by a SINGLE find-and-replace of original_value -> perturbed_value in the original condition. NO other characters may change:
   - Do NOT change articles (a/an) --- if the swap would require "a" -> "an" (e.g., "a multinomial" -> "an ordinal"), either pick a different perturbed_value that keeps the same article (e.g., "a binary"), OR include the article in both values (original_value: "an apocalyptic", perturbed_value: "a dystopian")
   - Similarly, if the swap affects a determiner or surrounding word, INCLUDE those words in both original_value and perturbed_value (e.g., original_value: "the same school", perturbed_value: "a different school")
   - Do NOT change apostrophes or punctuation (e.g., "worker's" must stay "worker's", not become "workers")
   - Do NOT rephrase surrounding words (e.g., "when they made" must not become "while they were finishing")
   - SELF-CHECK: mentally do find-and-replace of original_value -> perturbed_value in the original condition. If the result doesn't exactly match your perturbed_condition, fix it.
8. The perturbed value must be the SAME TYPE of attribute as the original --- both must describe the same dimension:
   - CORRECT: "short-lived" -> "long-running" (both describe duration)
   - CORRECT: "early" -> "late" (both describe temporal position --- NOT "in the middle of" which changes structure)
9. CRITICAL: Do NOT use synonyms, near-synonyms, or subset/superset values. The perturbed value must genuinely CONTRADICT the original when the original replaces it:
   - CORRECT: "popular" -> "controversial" (genuinely different attribute)
   - CORRECT: "African" -> "South American" (different region)
   - CORRECT: "music band" -> "solo artist" (genuinely different)
   TEST: Ask yourself --- if someone first hears the perturbed value, then hears the original value as a correction, would they recognize an actual CHANGE in factual content? If not, pick a different value.

Output JSON format:
{
  "perturbed_condition": "The full condition with the perturbed value",
  "original_value": "The original value/detail that was changed",
  "perturbed_value": "The new value/detail used as replacement",
  "reasoning": "Brief explanation of what was changed and why it's plausible"
}

<few-shot examples omitted for length>

Now generate a perturbed version of this condition:

Question: [[QUESTION]]
All Conditions (for context):
[[CONDITIONS]]
Target Condition to Perturb: [[TARGET_CONDITION]]

Output:
\end{Verbatim}

\begin{center}
\fcolorbox{black!30}{gray!8}{
  \begin{minipage}[t]{0.92\linewidth}\raggedright\footnotesize
  \textbf{Input} (function and argument 1 with identified swap span):\\
  \textbf{Function:}~What is the name of the ancient village?\\
  \textbf{Argument 1:}~The ancient village is part of a National Historic Landmark
  dating back to A.D.\ 1350.\\
  \textbf{Swap span:}~\texttt{A.D.\ 1350}
  \end{minipage}}
\end{center}
\begin{center}
\fcolorbox{black!30}{blue!4}{
  \begin{minipage}[t]{0.92\linewidth}\raggedright\footnotesize
  \textbf{Output} (two counterfactual candidates for argument 1):\\[2pt]
  \textbf{Counterfactual 1 (a):}~The ancient village is part of a National Historic
  Landmark dating back to \textbf{A.D.\ 1400}.~\textit{(span: \texttt{A.D.\ 1350}
  $\to$ \texttt{A.D.\ 1400})}\\[2pt]
  \textbf{Counterfactual 1 (b):}~The ancient village is part of a National Historic
  Landmark dating back to \textbf{A.D.\ 1300}.~\textit{(span: \texttt{A.D.\ 1350}
  $\to$ \texttt{A.D.\ 1300})}\\[3pt]
  \end{minipage}}
\end{center}

\clearpage
\subsection{Synthesize Prompts for Predecessor Functions}
\label{app:predecessor_prompt}

\begin{Verbatim}[fontsize=\scriptsize,frame=single,framesep=2mm,breaklines=true,breakanywhere=true,breaksymbolleft={},breaksymbolright={},baselinestretch=0.95]
You are designing a realistic multi-turn search conversation. You will generate a PREDECESSOR goal --- a question the user asked BEFORE moving on to the next goal.

=== SCENARIO ===

The user will eventually move on to asking: "[[NEXT_GOAL]]"
using these conditions (clues):
[[NEXT_CONDITIONS]]

Your task: Generate a PREDECESSOR goal --- what the user was asking about BEFORE they moved to the goal above.

=== KEY CONCEPT: CAUSAL DEPENDENCY ===

The predecessor and next goal must form a **causal chain**: the ANSWER to the predecessor question should naturally ENABLE or MOTIVATE the next question.

[[CHAIN_TYPE_INSTRUCTION]]

=== COMPLETE CONDITION SET ===

The predecessor goal must have ENOUGH conditions to be a meaningful, answerable search puzzle on its own:
1. Select a SUBSET of conditions from the next goal as "shared" conditions
2. Generate 1-3 NEW conditions specific to the predecessor goal
3. New conditions must provide enough context for an AI to attempt answering the predecessor
4. New conditions must describe the SAME underlying subject/domain as the shared conditions
5. New conditions must NOT contradict any shared conditions

=== RULES ===

- The predecessor must ask a GENUINELY DIFFERENT question than the next goal --- it must seek a DIFFERENT ENTITY or a DIFFERENT TYPE OF INFORMATION
- The predecessor must ask about a DIFFERENT TYPE OF INFORMATION than any goal in the avoid list
- CRITICAL: The ANSWER to the predecessor must be useful for or lead naturally to the next question
- Do NOT embed specific condition values (dates, numbers, names) into the question
- Keep the question short and natural --- like a real user typing in a chat
- The question must have a SPECIFIC, CONCRETE factual answer (not vague/open-ended)
- The predecessor question must NOT be answerable directly from the shared conditions themselves
- CRITICAL: The predecessor is the FIRST thing the user types in the conversation. It must be SELF-CONTAINED --- do NOT use phrases like "this author", "this series", "the same person", or any reference that assumes prior context. The question must make sense on its own without any preceding messages.
- CRITICAL: The predecessor must ask about a DIFFERENT ENTITY than the next goal. If the next goal asks "Who is X?", the predecessor must NOT also ask "Who is X?" with different clues. Instead, ask about a related entity Y whose answer reveals or leads to X.
[[FUTURE_CHAIN]]
[[AVOID_GOALS]]
[[SHARE_NUM_INSTRUCTION]]

=== OUTPUT FORMAT ===

Output valid JSON:
{
  "predecessor_goal": "The earlier question the user was asking (short and natural, SELF-CONTAINED with no dangling references, MAX 30 words)",
  "entity_sought": "The TYPE of entity the question seeks (e.g., 'person name', 'organization name', 'movie title', 'country', 'book title', 'year'). Must be 1-3 words.",
  "relevant_condition_ids": [list of integer condition IDs shared with the next goal],
  "new_conditions": ["New condition 1 specific to predecessor", "New condition 2 if needed"],
  "transition_reason": "Explain how the ANSWER to this predecessor naturally led the user to ask the next question",
  "transition_type": "One of: identify_then_seek, survey_then_focus, trace_then_follow, pivot_inquiry",
  "causal_link": "Explain specifically WHY the predecessor's answer is needed or useful for the next question",
  "reasoning": "Brief explanation of why this predecessor goal is natural and how the new conditions make it answerable"
}
\end{Verbatim}

\begin{center}
\fcolorbox{black!30}{gray!8}{
  \begin{minipage}[t]{0.92\linewidth}\raggedright\footnotesize
  \textbf{Input} (full intent: function and arguments):\\
  \textbf{Function:}~What is the name of the ancient village?\\
  \textbf{Arguments:}~the three arguments 1--3 listed above.
  \end{minipage}}
\end{center}
\begin{center}
\fcolorbox{black!30}{blue!4}{
  \begin{minipage}[t]{0.92\linewidth}\raggedright\footnotesize
  \textbf{Output} (three predecessor functions, one per taxonomy type):\\[2pt]
  \textbf{Predecessor (\emph{survey-then-focus}):}~What National Park
  Service--managed National Historic Landmarks near Santa Fe protect
  ancestral Pueblo sites?\\[1pt]
  \hspace*{1em}\textit{added arguments:} (i) NPS-managed federal heritage
  program; (ii) located in the Santa Fe area; (iii) publicly accessible for
  day visits with walking routes; (iv) focused on pre-contact ancestral
  Pueblo archaeology.\\[2pt]
  \textbf{Predecessor (\emph{trace-then-follow}):}~Which National Historic
  Landmark near Santa Fe preserves an ancient village with a walking
  trail?~\textit{(added arguments omitted)}\\[2pt]
  \textbf{Predecessor (\emph{pivot-inquiry}):}~How long is the interpretive
  trail at the National Historic Landmark near Santa Fe?~\textit{(added
  arguments omitted)}
  \end{minipage}}
\end{center}

\clearpage
\section{Additional Experiments and Analysis}
\label{app:more_exp}

\begin{figure}[t]
\centering

\begin{minipage}[t]{0.49\linewidth}
\vspace{0pt}
\centering
\includegraphics[width=\linewidth]{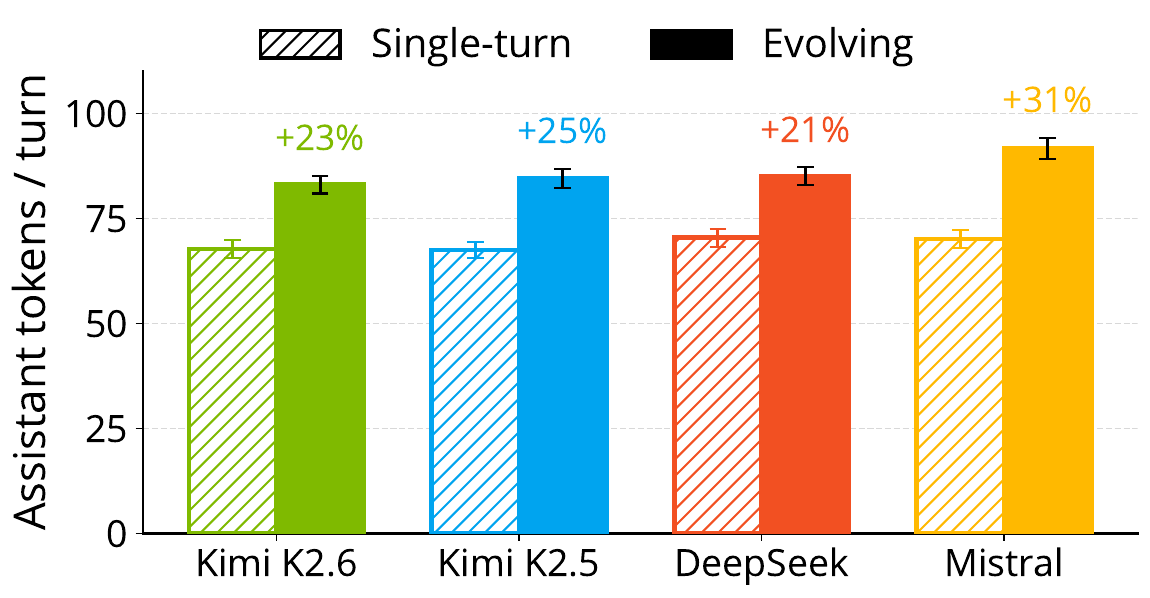}
\vspace{-0.3in}
\captionof{figure}{\textbf{Per-turn response length.} Average response length of open-source models across turns on BIRD-SQL. Later turns require models to spend more tokens reasoning over accumulated context.}
\label{fig:token_len}
\end{minipage}
\hfill
\begin{minipage}[t]{0.49\linewidth}
\vspace{0pt}
\centering
\includegraphics[width=\linewidth]{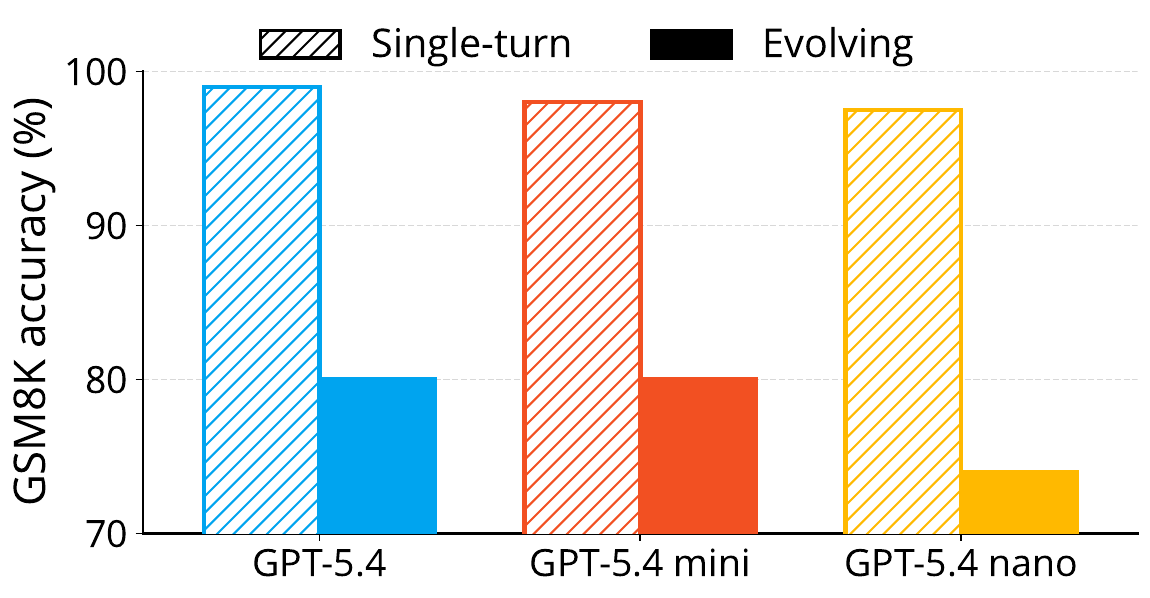}
\vspace{-0.3in}
\captionof{figure}{\textbf{Effect of model capacity.} Accuracy of the GPT 5.4 model family on GSM8K under single-turn and evolving-intent settings. Smaller models degrade more under evolving intent.}
\label{fig:model_capa}
\end{minipage}
\end{figure}

\subsection{Effect of Reasoning}

\begin{table}[t]
\centering
\small
\vspace{-.05in}
\caption{\textbf{Effect of reasoning.} Accuracy of GPT 5.1 on GSM8K and BIRD-SQL, comparing instant and reasoning modes under single-turn and evolving-intent settings.}
\label{tab:reasoning_ablation}
\vspace{-.1in}
\begin{tabular}{l cc cc}
\toprule
& \multicolumn{2}{c}{\textbf{GSM8K}} & \multicolumn{2}{c}{\textbf{BIRD-SQL}} \\
\cmidrule(lr){2-3} \cmidrule(lr){4-5}
\textbf{Model} & Single & Evolve & Single & Evolve \\
\midrule
GPT-5.1 (instant) & 97.5 & 83.0 & 73.0 & 64.0 \\
\quad + reasoning & 98.0 & 82.0 & 72.0 & 66.0 \\
\bottomrule
\end{tabular}
\vspace{-.1in}
\end{table}

To examine whether stronger per-turn reasoning mitigates performance degradation, we compare GPT 5.1 in its instant and reasoning modes on GSM8K and BIRD-SQL. We choose these two datasets because their final answers are well defined and automatically verifiable, while the tasks do not require long-horizon tool interaction or open-ended planning, allowing us to isolate whether additional inference-time reasoning helps agents adapt to evolving intent. As shown in \Cref{tab:reasoning_ablation}, reasoning yields no meaningful improvement: it only marginally changes single-turn accuracy and provides at most a small gain on BIRD-SQL under the evolving-intent setting, while GSM8K slightly decreases. This is consistent with prior observations that stronger reasoning does not necessarily resolve failures caused by under-specified user intent~\citep{laban2026llmsa}. We cautiously conjecture that this is because LLMs are largely trained to use the given context to predict the correct answer, whereas our setting often requires the model to selectively discard, revise, or override earlier context as the user's intent evolves. Thus, the bottleneck is not simply local reasoning ability, but maintaining an up-to-date belief over the user's intent across turns. 

\subsection{Response Length Analysis} 

We further analyze how response length changes under evolving intent using open-source models whose response traces are observable; for closed-source models, hidden reasoning traces make such estimation difficult. On BIRD-SQL, we compare four open-source models and measure their average response length in the single-turn and evolving-intent settings. As shown in \Cref{fig:token_len}, models produce longer responses under evolving intent than in the single-turn setting. This suggests that evolving intent requires models to spend more tokens processing accumulated and potentially revised context. 

\subsection{Effect of Model Capacity} 

We additionally examine the effect of model capacity within the GPT 5.4 model family. Although the exact model sizes are not disclosed, the model provider describes GPT 5.4 mini and nano as smaller variants of GPT 5.4; we therefore treat them as capacity variants and evaluate all three models in reasoning mode. On GSM8K, the three variants achieve nearly identical accuracy in the single-turn setting. However, under the evolving-intent setting, GPT 5.4 and GPT 5.4 mini remain comparable, whereas GPT 5.4 nano shows a substantially larger degradation. This suggests that evolving-intent interaction may require additional model capacity, even when single-turn performance appears saturated.

\begin{table}[t]
\caption{
\textbf{Effect of the number of turns.} Accuracy of GPT-5.1 on GSM8K under varying conversation lengths and intent-transition compositions.
Increasing the conversation from four to seven turns through intent-consistent repetition does not reduce accuracy, whereas adding intent transitions at the same seven-turn length reduces accuracy.
}
\vspace{-0.1in}
\label{tab:num_turns}
\centering
\small
\begin{tabular}{lclc}
\toprule
\textbf{Setting} & \textbf{\# Turns} & \textbf{Transition Composition} & \textbf{Accuracy (\%)} \\
\midrule
Single-turn
    & 1
    & -- 
    & 98.0 \\
Multi-turn
    & 4
    & 1 init + 1 reveal + 1 switch + 1 change
    & 87.0 \\
Multi-turn (w/ repeats)
    & 7
    & 1 init + 1 reveal + 1 switch + 1 change + 3 repeats
    & 87.5 \\
Multi-turn (more transitions)
    & 7
    & 1 init + 2 reveals + 2 switches + 2 changes
    & 82.0 \\
\bottomrule
\end{tabular}
\end{table}

\subsection{Effect of the Number of Conversation Turns}

To distinguish the effect of evolving intent from that of conversation turn (i.e., number of user turn), we construct turn-matched controls that vary the number of turns while controlling the underlying intent dynamics. To this end, we repeat the turn (of the previous conversation), making no intent change while increasing the number of turns. Here, as shown in Table \ref{tab:num_turns}, we found that adding more turns without intent changes does not reduce performance, i.e., accuracy remains comparable to the original four-turn setting. In contrast, replacing these repeated turns with additional intent transitions reduces accuracy. These results indicate that the performance degradation is driven primarily by changes in user intent, rather than by the number of turns alone.

\begin{table}[t]
\caption{
\textbf{RL with evolving intent.} Performance of Qwen3-4B before and after training on evolving-intent examples constructed using our framework. We train with GRPO on the transformed GSM8K examples using standard outcome reward. Training improves evolving-intent performance while preserving single-turn performance.
}
\vspace{-0.1in}
\label{tab:evolving-intent-training}
\centering
\small
\begin{tabular}{lcc}
\toprule
\textbf{Model} & \textbf{Single-Turn} & \textbf{Evolving Intent} \\
\midrule
Qwen3-4B & 94.0 & 64.0 \\
RL on Evolving Intent & 95.0 & 76.0 \\
\bottomrule
\end{tabular}
\end{table}

\subsection{RL with Evolving Intent}

The main experiments focus on evaluating frontier models and analyzing their failure modes under evolving user intent. However, we remark that our framework can also be used to construct training examples by transforming existing single-turn tasks into multi-turn verifiable conversations. This provides a direct way to train models to track and act on changing user intent. As a preliminary demonstration, we train Qwen3-4B for fewer than 50 steps on the evolving-intent version of GSM8K using GRPO \cite{shao2024deepseekmath} with a standard outcome reward (i.e., measured in the final turn of the conversation and treat all conversation is correct if the reward is positive). As shown in Table \ref{tab:evolving-intent-training}, training improves evolving-intent performance from 64.0 to 76.0 while preserving single-turn performance.

\end{document}